\pdfoutput=1
\documentclass[10pt,twocolumn,letterpaper]{article}

\usepackage{template/cvpr}
\usepackage{graphicx}
\usepackage{amsmath}
\usepackage{amssymb}
\usepackage{booktabs}
\usepackage{enumitem}

\usepackage{caption}
\usepackage{subcaption}
\usepackage{cuted}
\usepackage{makecell}
\usepackage{adjustbox}

\usepackage[dvipsnames]{xcolor}
\definecolor{cvprblue}{rgb}{0.21,0.49,0.74}
\usepackage{hyperref}
\hypersetup{breaklinks,colorlinks,citecolor=cvprblue}

\usepackage[capitalize]{cleveref}
\crefname{section}{Sec.}{Secs.}
\Crefname{section}{Section}{Sections}
\Crefname{table}{Table}{Tables}
\crefname{table}{Tab.}{Tabs.}
\newcommand{\authorsep}{\hspace{8pt}}
\newcommand{\affiliationsep}{\hspace{8pt}}
\newcommand{\EqCont}{${}^{*}$}
\newcommand{\AffI}{${}^1$}

\newcommand{\AffIandII}{${}^{1,2}$}
\newcommand{\Aff}{${}^{1,3}$}

\usepackage{url}

\title{HD-Painter: High-Resolution and Prompt-Faithful \\Text-Guided Image Inpainting with Diffusion Models}

\author{Hayk Manukyan\AffI\EqCont \authorsep
Andranik Sargsyan\AffI\EqCont \authorsep
Barsegh Atanyan\AffI \authorsep
Zhangyang Wang\AffIandII \\
Shant Navasardyan\AffI \authorsep
Humphrey Shi\Aff \\
\small${}^1$Picsart AI Research (PAIR) \affiliationsep
\small${}^2$UT Austin \affiliationsep
\small${}^3$Georgia Tech
\\{\small \textbf{\url{https://github.com/Picsart-AI-Research/HD-Painter}}}
}

\newtheorem{prop}{{\sc Claim}}

\begin{document}
\maketitle

\begin{strip}
\vspace{-15mm}
    \centering
    \includegraphics[width=\linewidth]{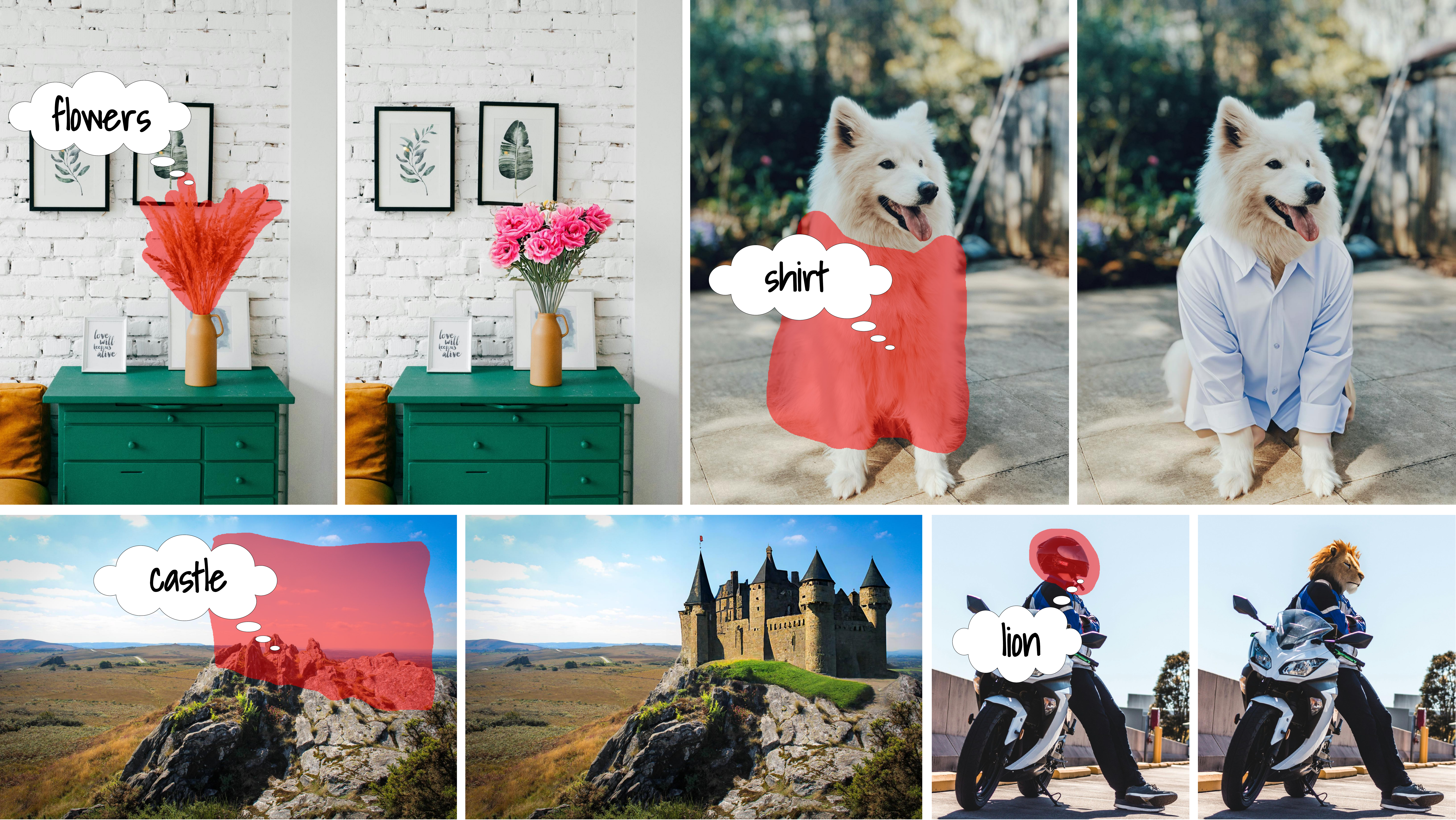}
    \captionof{figure}{High-resolution (the large side is $2048$ in all these examples) text-guided image inpainting results with our approach.
    The method is able to faithfully fill the masked region according to the prompt even if the combination of the prompt and the known region is highly unlikely. Zoom in to view high-resolution details.
    }
    \label{fig:teaser-image}
\end{strip}

\def\thefootnote{*}\footnotetext{Equal contribution.}\def\thefootnote{\arabic{footnote}}

\begin{abstract}
Recent progress in text-guided image inpainting, based on the unprecedented success of text-to-image diffusion models, has led to exceptionally realistic and visually plausible results.
However, there is still significant potential for improvement in current text-to-image inpainting models, particularly in better aligning the inpainted area with user prompts and performing high-resolution inpainting.
Therefore, we introduce \textit{HD-Painter}, a \textbf{training-free} approach that \textbf{accurately follows prompts} and coherently \textbf{scales to high resolution} image inpainting.
To this end, we design the \textit{Prompt-Aware Introverted Attention (PAIntA)} layer enhancing self-attention scores by prompt information resulting in better text aligned generations.
To further improve the prompt coherence we introduce the \textit{Reweighting Attention Score Guidance (RASG)} mechanism seamlessly integrating a post-hoc sampling strategy into the general form of DDIM to prevent out-of-distribution latent shifts.
Moreover, HD-Painter allows extension to larger scales by introducing a specialized super-resolution technique customized for inpainting, enabling the completion of missing regions in images of up to 2K resolution. 
Our experiments demonstrate that HD-Painter surpasses existing state-of-the-art approaches quantitatively and qualitatively across multiple metrics and a user study. 
Code is publicly available at: \textbf{\url{https://github.com/Picsart-AI-Research/HD-Painter}}
\end{abstract}.

\section{Introduction}

The recent wave of diffusion models \cite{DDPM,DDIM} has taken the world by storm, becoming an increasingly integral part of our everyday lives.
After the unprecedented success of text-to-image models \cite{StableDiffusion,DALLE2,Imagen,wu2022nuwa} diffusion-based image manipulations such as prompt-conditioned editing \cite{prompt2prompt,brooks2023instructpix2pix}, controllable generation \cite{ControlNet,t2i_adapter}, personalized and specialized image synthesis \cite{ruiz2023dreambooth,textualinversion,lu2023specialist} became hot topics in computer vision leading to a huge amount of applications.
Particularly, text-guided image completion or inpainting  \cite{wang2023imageneditor,wu2022nuwa,avrahami2022blendeddiffusion} allows users to generate new content in user-specified regions of given images based on textual prompts (see Fig. \ref{fig:teaser-image}), 
leading to use cases like retouching specific areas of an image, replacing or adding objects, and modifying subject attributes such as clothes, colors, or emotion.

Pretrained text-to-image generation models such as Stable Diffusion \cite{StableDiffusion}, Imagen \cite{Imagen}, and Dall-E 2 \cite{DALLE2} can be adapted for image completion by blending diffused known regions with generated (denoised) unknown regions during the backward diffusion process. 
Although such approaches \cite{avrahami2022blendeddiffusion,BlendedLatentDiffusion} produce visually plausible completions, they are not well harmonized and lack global scene understanding, especially when denoising in high diffusion timesteps.

To address this, existing methods \cite{StableDiffusion,GLIDE,podell2023sdxl,Imagen}, modify pretrained text-to-image models to take additional context information and fine-tune specifically for text-guided image completion. GLIDE \cite{GLIDE} and Stable Inpainting \cite{StableDiffusion} concatenate the mask and the masked image as additional channels to the input of the diffusion UNet, initializing the new convolutional weights with zeros, then fine tune the modified model using random masks together with the initial prompt.

However, SmartBrush \cite{SmartBrush} and Imagen Editor \cite{wang2023imageneditor} 
mention the weak image-text alignment of such models, 
attributing it to the random masking strategies, 
and the misalignment of the global prompts used during training with the local 
context of the masked region.
In this paper, we will address this issue as \textit{prompt neglect}.
To alleviate this problem, both papers introduce novel, object-aware masking strategies.
Additionally SmartBrush proposes BLIP captioning approach, 
to ensure a better alignment of the inpainting prompt with the masked region.
Nonetheless, we find that while this approach reduces the amount of prompt neglect, 
it also decreases the generation quality \cref{tab:metrics}.

We noticed that prompt neglect is commonly expressed in two ways:
either the model fills in the masked region with background (\textit{background dominance}, \cref{fig:qualitative_comparison}, columns 1, 3, 5), 
or the model completes a nearby object partially occluded by the mask (\textit{nearby object dominance}, \cref{fig:qualitative_comparison}, columns 2, 4, 6).
In both cases the issue seems to be caused by the model preferring the local context 
of the known region to the textual information provided by the prompt.

To address the mentioned problems we introduce \textit{Prompt-Aware Introverted Attention (PAIntA)} block without any \textit{training or fine-tuning} requirements. 
PAIntA enhances the self-attention scores according to the given textual condition aiming to decrease the impact of non-prompt-relevant information from the image known region while increasing the contribution of the prompt-aligned known pixels.

To improve the text-alignment of the generation results even further we apply a \textit{post-hoc guidance} mechanism by leveraging the cross-attention scores. 
However the vanilla post-hoc guidance mechanism used by seminal works such as \cite{DiffusionBeatGAN, self_guidance}, etc. may lead to generation quality degradation due to out-of-distribution shifts caused by the additional gradient term in the backward diffusion equation (see \cref{eq:ddim_deterministic_process}).
To this end we propose \textit{Reweighting Attention Score Guidance (RASG)}, a post-hoc mechanism seamlessly integrating the gradient component in the general form of DDIM process.
This allows to simultaneously guide the sampling towards more prompt-aligned latents and keep them in their trained domain leading to visually plausible inpainting results.

With the combination of PAIntA and RASG our method gains a significant advantage over the current state-of-the-art approaches by solving the issue of prompt neglect. In addition, by leveraging high-resolution diffusion models and time-iterative blending technology we design a simple yet effective pipeline for up to $2048\times 2048$ resolution inpainting.

To summarize, our main contributions are as follows:
\begin{itemize} 
  \item We introduce the \textit{Prompt-Aware Introverted Attention (PAIntA)} layer to alleviate the prompt neglect issues of background and nearby object dominance in text-guided image inpainting.
  \item To further improve the text-alignment of generation we present the \textit{Reweighting Attention Score Guidance (RASG)} strategy which enables to prevent out-of-distribution shifts while performing post-hoc guided sampling.
  \item Our designed pipeline for text-guided image completion is \textit{training-free} and demonstrates a significant advantage over current state-of-the-art approaches quantitatively and qualitatively. 
  Moreover, with the additional help of our simple yet effective inpainting-specialized super-resolution framework we make high-resolution (up to $2048\times 2048$) image completion possible.
\end{itemize}

\section{Related Work}

\subsection{Image Inpainting}
 
Image inpainting is the task of filling missing regions of the image in a visually plausible manner. 
Early deep learning approaches such as \cite{yu2018generative,yi2020contextual,navasardyan2020image} introduce mechanisms to propagate deep features from known regions. 
Later \cite{zhao2021large,zheng2022cmgan,xu2023image,MI-GAN} utilize StyleGAN-v2-like \cite{karras2020analyzing} decoder and discriminative training for better image detail generation. 

Image inpainting also benefited from diffusion models, particularly with the emergence of text-guided inpainting.
Given a pre-trained text-to-image diffusion model \cite{avrahami2022blendeddiffusion,BlendedLatentDiffusion} 
replace the unmasked region of the latent by the noised version of the known region during sampling. 
However, as noted by \cite{GLIDE}, this leads to poor generation quality, 
as the denoising network only sees the noised version of the known region. 
\cite{GLIDE, wang2023imageneditor, podell2023sdxl, SmartBrush} fine-tune pretrained text-to-image models 
for text-guided image inpainting by conditioning the denoising model on the inpainting mask and the known region, concatenating them with the input latents.
\cite{SmartBrush,wang2023imageneditor}, in particular, use object-aware masking strategies, to improve image-text alignment of training samples.
Alternatively, \cite{ControlNet} obtains an inpainting model by attaching trainable modules to the UNet, while keeping the base model unchanged.
We propose a training-free approach leveraging plug-and-play components PAIntA and RASG, improving text-prompt alignment. 
Moreover, our approach allows inpainting on high-resolution images (up to $2048\times 2048$). 

\subsection{Inpainting-Specific Architectural Blocks}
Early deep learning approaches were designing special layers for better/more efficient inpainting. 
Particularly, \cite{PConv,GatedConv,navasardyan2020image} introduce special convolutional layers dealing with the known region of the image to effectively extract the information useful for visually plausible image completion. 
\cite{yi2020contextual} introduces the contextual attention layer reducing the unnecessarily heavy computations of all-to-all self-attention for high-quality inpainting.
In this work we propose Prompt-Aware Introverted Attention (PAIntA) layer, specifically designed for text-guided image inpainting. 
It aims to decrease (increase) the prompt-irrelevant (-relevant) information from the known region for better text aligned inpainting generation.

\subsection{Post-Hoc Guidance in Diffusion Process}
Post-hoc guidance methods are backward diffusion sampling techniques which guide the next step latent prediction towards a specific objective function minimization. Such approaches appear to be extremely helpful when generating visual content especially with an additional constraint.
Particularly \cite{DiffusionBeatGAN} introduced classifier-guidance aiming to generate images of a specific class. 
Later CLIP-guidance was introduced by \cite{GLIDE} leveraging CLIP \cite{CLIP} as an open-vocabulary classification method.
LDM \cite{StableDiffusion} further extends the concept to guide the diffusion sampling process by any image-to-image translation method, particularly guiding a low-resolution trained model to generate $\times 2$ larger images.
\cite{AttendExcite} guides image generation by maximizing the maximal cross-attention score relying on multi-iterative optimization process resulting in more text aligned results.
\cite{self_guidance} goes even further by utilizing the cross-attention scores for object position, size, shape, and appearance guidances.
All the mentioned post-hoc guidance methods shift the latent generation process by a gradient term (see \cref{eq:shifting_gradient_term}) sometimes leading to image quality degradations.

To this end we propose the Reweighting Attention Score Guidance (RASG) mechanism allowing to perform post-hoc guidance with any objective function \textbf{while preserving the diffusion latent domain}.
Specifically for inpainting task, to alleviate the issue of prompt neglect, we benefit from a guidance objective function based on the open-vocabulary segmentation properties of cross-attentions.

\section{Method}

We first formulate the text-guided image completion problem followed by an introduction to diffusion models, particularly Stable Diffusion (\cite{StableDiffusion}) and Stable Inpainting.
We then discuss the overview of our method and its components.
Afterwards we present our Prompt-Aware Introverted Attention (PAIntA) block and Reweighting Attention Score Guidance (RASG) mechanism in detail.
Lastly our inpainting-specific super-resolution technique is introduced.

Let $I\in\mathbb{R}^{H\times W\times 3}$ be an RGB image, $M\in\{0,1\}^{H\times W}$ be a binary mask indicating the region in $I$ one wants to inpaint with a textual prompt $\tau$.
The goal of text-guided image inpainting is to output an image $I^c\in\mathbb{R}^{H\times W\times 3}$ such that $I^c$ contains the objects described by the prompt $\tau$ in the region $M$ while outside $M$ it coincides with $I$, i.e. $I^c\odot(1-M) = I\odot(1-M)$.

\subsection{Stable Diffusion and Stable Inpainting}

Stable Diffusion (SD) is a diffusion model that functions within the latent space of an autoencoder $\mathcal{D}(\mathcal{E}(\cdot))$ (VQ-GAN \cite{VQGAN_paper} or VQ-VAE \cite{VQVAE_paper}) where $\mathcal{E}$ denotes the encoder and $\mathcal{D}$ the corresponding decoder.
Specifically, let $I\in\mathbb{R}^{H\times W\times 3}$ be an image and $x_0 = \mathcal{E}(I)$, consider the following forward diffusion process with hyperparameters $\{\beta_t\}_{t=1}^{T}\subset [0,1]$:
\begin{equation}
    q(x_t|x_{t-1}) = \mathcal{N}(x_t; \sqrt{1-\beta_t}x_{t-1}, \beta_t I), \;
    t = 1,..,T
\end{equation}
where $q(x_t|x_{t-1})$ is the conditional density of $x_t$ given $x_{t-1}$, and $\{x_t\}_{t=0}^T$ is a Markov chain.
Here $T$ is large enough to allow an assumption $x_T \sim \mathcal{N}(\textbf{0},\textbf{1})$. 
Then SD learns a backward process (below similarly, $\{x_t\}_{t=T}^0$ is a Markov chain)
\begin{equation}
\label{eq:diff_backward_markov_chain}
    p_\theta(x_{t-1}|x_t) = \mathcal{N}(x_{t-1};\mu_\theta(x_t,t),\sigma_t\textbf{1}), \;
    t=T,..,1,
\end{equation}
and hyperparameters $\{\sigma_t\}_{t=1}^T$, allowing the generation of a signal $x_0$ from the standard Gaussian noise $x_T$.
Here $\mu_{\theta}(x_t,t)$ is defined by the predicted noise $\epsilon_{\theta}^t(x_t)$ modeled as a neural network (see \cite{DDPM}): 
$
\mu_{\theta}(x_t,t) = \frac{1}{\sqrt{\beta_t}}\left(x_t - \frac{\beta_t}{\sqrt{1-\alpha_t}}\epsilon_{\theta}^t(x_t)\right)
$.
Then $\hat{I} = \mathcal{D}(x_0)$ is returned.

The following claim can be derived from the main DDIM principle, Theorem 1 in \cite{DDIM}.

\begin{prop}
\label{prop:DDIM_sigma_proposition}
    After training the diffusion backward process (Eq. \ref{eq:diff_backward_markov_chain}) the following $\{\sigma_t\}_{t=1}^T$-parametrized family of DDIM sampling processes can be applied to generate high-quality images:
    \begin{equation}
    \begin{aligned}
    \label{eq:ddim_stochastic_process}
        x_{t-1} = \sqrt{\alpha_{t-1}} \frac{x_t - \sqrt{1 - \alpha_t}\epsilon^t_\theta(x_t)}{\sqrt{\alpha_t}} + \\
        \sqrt{1 - \alpha_{t-1} - \sigma_t ^ 2} \epsilon^t_\theta(x_t) +  \sigma_t \epsilon_t,
    \end{aligned}
    \end{equation}
    where $\epsilon_t\sim\mathcal{N}(\textbf{0},\textbf{1})$, $\alpha_t = \prod_{i=1}^{t}(1-\beta_i)$, and $0\leq\sigma_t\leq \sqrt{1-\alpha_{t-1}}$ can be arbitrary parameters.
\end{prop}

Usually (e.g. in SD or Stable Inpainting described below) $\sigma_t=0$ is taken to get a deterministic process:
\begin{equation}
\begin{aligned}
\label{eq:ddim_deterministic_process}
    x_{t-1} = \sqrt{\alpha_{t-1}}\left(\frac{x_t - \sqrt{1-\alpha_t}\epsilon^t_{\theta}(x_t)}{\sqrt{\alpha_t}}\right) + \\
    \sqrt{1-\alpha_{t-1}}\epsilon^t_{\theta}(x_t), \; t=T,\ldots,1.
\end{aligned}
\end{equation}
For text-to-image synthesis, SD guides the processes with a textual prompt $\tau$.
Hence the function $\epsilon^t_{\theta}(x_t) = \epsilon^t_{\theta}(x_t,\tau)$, modeled by a UNet-like (\cite{UNet_paper}) architecture, is also conditioned on $\tau$ by its cross-attention layers.
For simplicity sometimes we skip $\tau$ in writing $\epsilon^t_{\theta}(x_t,\tau)$.

As mentioned earlier, Stable DIffusion can be modified and fine-tuned for text-guided image inpainting.
To do so \cite{StableDiffusion} concatenate the features of the masked image $I^M = I\odot (1-M)$ obtained by the encoder $\mathcal{E}$, and the (downscaled) binary mask $M$ to the latents $x_t$ and feed the resulting tensor to the UNet to get the estimated noise
$\epsilon^t_{\theta}([x_t, \mathcal{E}(I^M),down(M)], \tau)$,
where $down$ is the downscaling operation to match the shape of the latent $x_t$. 
Newly added convolutional filters are initialized with zeros while the rest of the UNet from a pretrained checkpoint of Stable Diffusion.
Training is done by randomly masking images and optimizing the model to reconstruct them based on image captions from the LAION-5B (\cite{schuhmann2022laion}) dataset.
The resulting model shows visually plausible image completion and we refer to it as \textit{Stable Inpainting}.

\subsection{HD-Painter: Overview}

\begin{figure*}[h!]
    \centering
    \includegraphics[width=\textwidth]{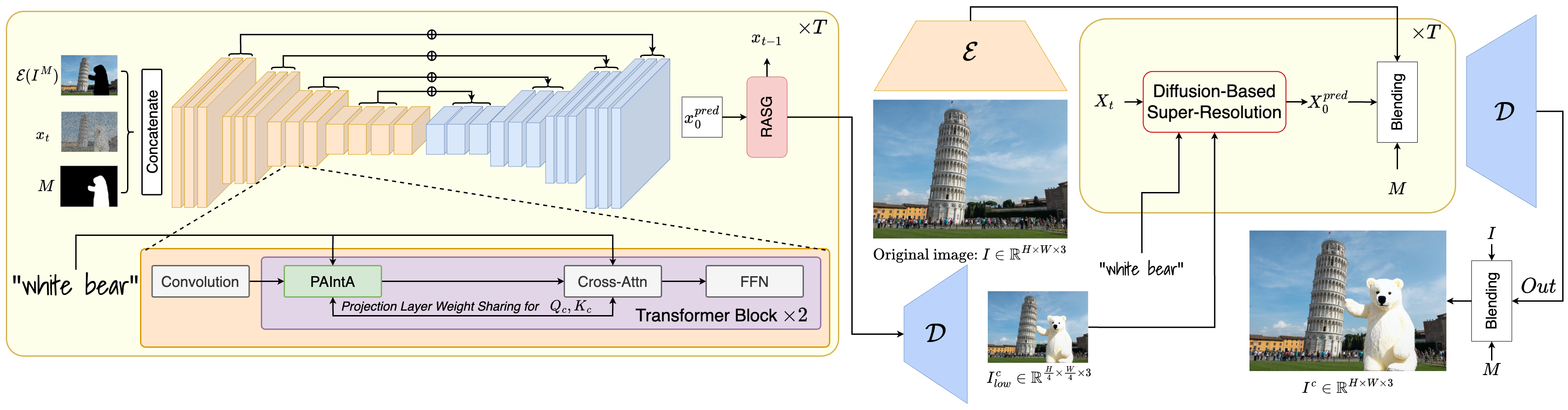}
    \caption{
    Our method has two stages: image completiton, and inpainting-specialized super-resolution ($\times 4$). For image completion in each diffusion step we denoise the latent $x_t$ by conditioning on the inpainting mask $M$ and the masked downscaled image $I^M = down(I)\odot(1-M)\in\mathbb{R}^{\frac{H}{4}\times \frac{W}{4}\times 3}$ (encoded with the VAE encoder $\mathcal{E}$). To make better alignement with the given prompt our \textit{PAIntA} block is applied instead of self-attention layers.
    After predicting the denoised $x^{pred}_0$ in each step $t$, we provide it to our \textit{RASG} guidance mechanism to estimate the next latent $x_{t-1}$. 
    For inpainting-specific super resolution we condition the high-resolution latent $X_t$ denoising process by the lower resolution inpainted result $I^c_{low}$, followed by blending $X^{pred}_0\odot M + \mathcal{E}(I)\odot(1-M)$. Finally we get $I^c$ by Poisson blending the decoded output with the original image $I$.
    }
    \label{figures:method_arch}
\end{figure*}

The overview of our method is presented in Fig. \ref{figures:method_arch}. 
The proposed pipeline is composed of two stages: text-guided image inpainting on the resolution $H/4\times W/4$ is applied followed by the inpainting-specific $\times4$ super-resolution of the generated content.

To complete the missing region $M$ according to the given prompt $\tau$ we take a pre-trained inpainting diffusion model like Stable Inpainting, replace the self-attention layers by PAIntA layers, and perform a diffusion backward process by applying our RASG mechanism.
After getting the final estimated latent $x_0$, it is decoded resulting in an inpainted image $I^{c}_{low} = \mathcal{D}(x_0)\in\mathbb{R}^{\frac{H}{4}\times\frac{W}{4}}$.

To inpaint the original size image $I\in\mathbb{R}^{H\times W}$ we utilize the super-resolution stable diffusion from \cite{StableDiffusion}.
We apply the diffusion backward process of SD starting from $X_T\sim\mathcal{N(\textbf{0},\textbf{1})}$ and conditioned on the low resolution inpainted image $I^c_{low}$. 
After each step we blend the denoised $X^{pred}_0$ with the original image's encoding $\mathcal{E}(I)$ in the known region indicated by the mask $(1-M)\in \{0,1\}^{H\times W}$ and derive the next latent $X_{t-1}$ by Eq. \ref{eq:ddim_deterministic_process}. 
After the final step we decode the latent by $\mathcal{D}(X_0)$ and use Poisson blending (\cite{poisson_blending}) with $I$ to avoid edge artifacts.

\subsection{Prompt-Aware Introverted Attention (PAIntA)}

\begin{figure*}
    \centering
    \includegraphics[width=0.92\linewidth]{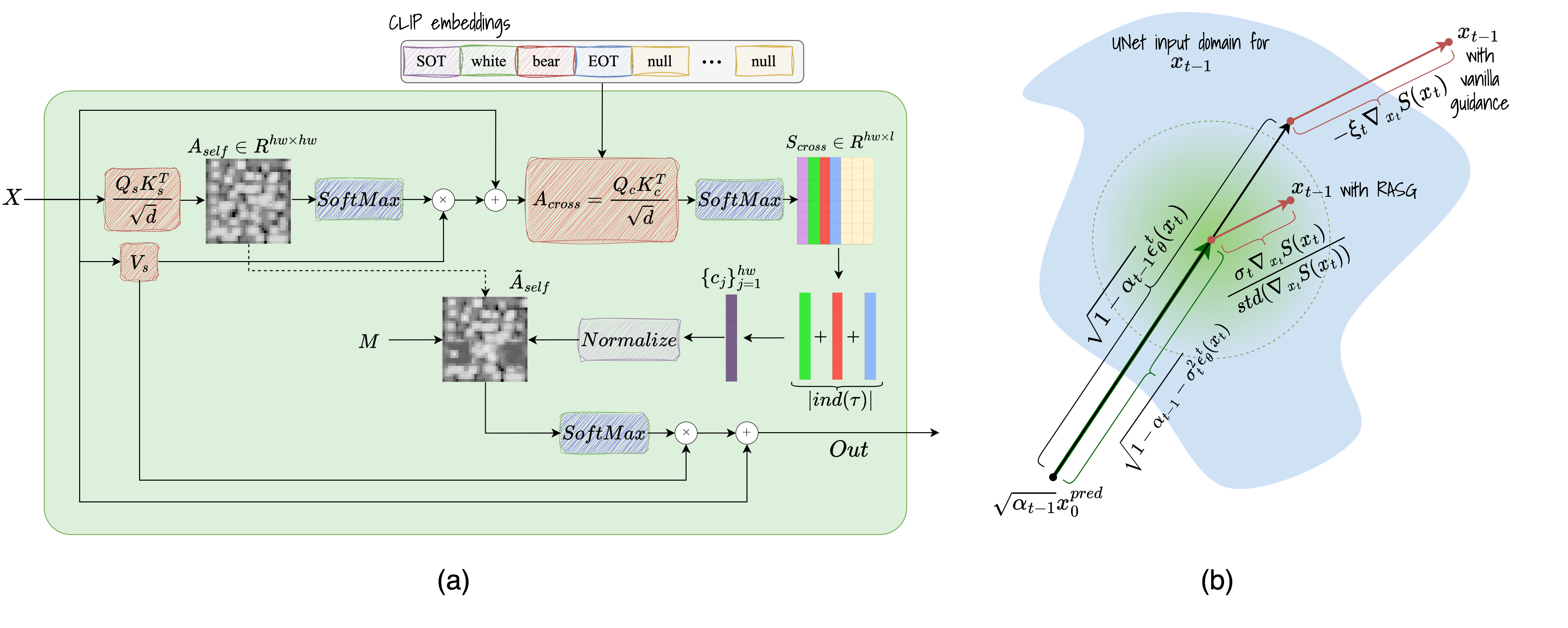}
    \caption{
    (a) PAIntA block takes an input tensor $X\in\mathbb{R}^{h\times w\times 3}$ and the CLIP embeddings of $\tau$. After computing the self- and cross-attention scores $A_{self}$ and $A_{cross}$, we update the former (Eq. \ref{eq:self_attn_update}) by scaling with the normalized values $\{c_j\}_{j=1}^{hw}$ obtained from $S_{cross} = SoftMax(A_{cross})$. Finally the the updated attention scores $\tilde{A}_{self}$ are used for the convex combination of the values $V_s$ to get the residual of PAIntA's output.
    (b) RASG mechanism takes the predicted scaled denoised latent $\sqrt{\alpha_{t-1}}x^{pred}_0 = \frac{\sqrt{\alpha_{t-1}}}{\sqrt{\alpha_t}}\left(x_t - \sqrt{1-\alpha_t}\epsilon_{\theta}(x_t)\right)$ and guides the $x_{t-1}$ estimation process towards minimization of $S(x_t)$ defined by Eq. \ref{eq:defining_function_S}. 
    Gradient reweighting makes the gradient term close to being sampled from $\mathcal{N}(\textbf{0},\textbf{1})$ (green area) by so ensuring the domain preservation (blue area).
    }
    \label{fig:RASG_domain_fig}
\end{figure*}

Throughout our experiments we noticed that existing approaches, such as Stable Inpainting, tend to ignore the user-provided prompt relying more on the visual context around the inpainting area. 
In the introdution we categorized this issue into two classes based on user experience: \textit{background dominance} and \textit{nearby object dominance}.
Indeed, for example in Fig. \ref{fig:qualitative_comparison}, rows 1, 3, 4, the existing solutions (besides BLD) fill the region with background, and in rows 5, 6, they prefer to continue the animal and the car instead of generating a boat and flames respectively.
We hypothesize that the \textit{visual context dominance} over the prompt is attributed to the \textit{prompt-free, only-spatial} nature of self-attention layers. 
To support this we visualize the self-attention scores (see Appendix) and observe a high similarity between the inpainted tokens and such known tokens of the image which have low similarity with the prompt (for more details see Appendix).
Therefore, to alleviate the issue, we introduce a plug-in replacement for self-attention, Prompt-Aware Introverted Attention (PAIntA, see Fig. \ref{fig:RASG_domain_fig} (a)) which utilizes the inpainting mask $M$ and cross-attention matrices to control the self-attention output in the unknown region.
Below we discuss PAIntA in detail.

Let $X\in\mathbb{R}^{(h\times w)\times d}$ be the input tensor of PAIntA.
Similar to self-attention, PAIntA first applies projection layers to get the queries, keys, and values we denote by $Q_s, K_s, V_s \in \mathbb{R}^{(h\times w)\times d}$ respectively, and the similarity matrix
$A_{self} = \frac{Q_sK^T_s}{\sqrt{d}} \in \mathbb{R}^{hw\times hw}$.
Then we mitigate the too strong influence of the known region over the unknown by adjusting the attention scores of known pixels contributing to the inpainted region.
Specifically, leveraging the prompt $\tau$, PAIntA defines a new similarity matrix:
\begin{equation}
\begin{aligned}
    \tilde{A}_{self}\in\mathbb{R}^{hw\times hw}, \quad \\
    (\Tilde{A}_{self})_{ij} = 
    \begin{cases}
        c_j \cdot (A_{self})_{ij} & M_i = 1 \; \mbox{and} \; M_j = 0, \\
        (A_{self})_{ij} & \mbox{otherwise},
    \end{cases}
    \label{eq:self_attn_update}
\end{aligned}
\end{equation}
where $c_j$ shows the alignment of the $j^{th}$ feature token (pixel) with the given textual prompt $\tau$.

We define $\{c_j\}_{j=1}^{hw}$ using the cross-attention spatio-textual similarity matrix 
$
S_{cross} = SoftMax(Q_{c}K_{c}^T/\sqrt{d})
$,
where $Q_c \in \mathbb{R}^{(h \times w) \times d}$, $K_c \in \mathbb{R}^{l \times d}$ are query and key tensors of corresponding cross-attention layers, and $l$ is the number of tokens of the prompt $\tau$.
Specifically, we consider CLIP text embeddings of the prompt $\tau$ and separate the ones which correspond to the words of $\tau$ and \textit{End of Text} (EOT) token (in essence we just disregard the SOT token and the null-token embeddings), and denote the set of chosen indices by $ind(\tau)\subset \{1,2,\ldots,l\}$. 
We include EOT since (in contrast with SOT) it contains information about the prompt $\tau$ according to the architecture of CLIP text encoder.
For each $j^{th}$ pixel we define its similarity with the prompt $\tau$ by summing up it's similarity scores with the embeddings indexed from $ind(\tau)$, i.e. $c_j = \sum_{k\in ind(\tau)} (S_{cross})_{jk}$.
Also, we found beneficial to normalize the scores
$
c_j = clip\left(\frac{c_j - median(c_k;\; k=1,\ldots,hw)}{max(c_k;\; k=1,\ldots,hw)}, 0, 1\right)
$,
where $clip$ is the clipping operation between $[0,1]$.

Note that in vanilla SD cross-attention layers come after self-attention layers, hence in PAIntA to get query and key tensors $Q_c, K_c$ we borrow the projection layer weights from the next cross-attention module (see Fig. \ref{figures:method_arch}).
Finally we get the output of the PAIntA layer with the residual connection with the input: 
$
Out = X + SoftMax(\Tilde{A}_{self})\cdot V_s.
$

\subsection{Reweighting Attention Score Guidance (RASG)}

To further enhance the generation alignment with the prompt $\tau$ we adopt a post-hoc sampling guidance mechanism \cite{DiffusionBeatGAN} with an objective function $S(x)$ leveraging the open-vocabulary segmentation properties of cross-attention layers.
Specifically\footnote{for brevity: $\epsilon^t_{\theta}(x_t) = \epsilon^t_{\theta}([x_t,\mathcal{E}(I^M), down(M)],\tau)$.} at each step the following update rule is used after predicting the noise $\epsilon_{\theta}^t(x_t)$:
$\hat{\epsilon}_{\theta}^t(x_t) \leftarrow \epsilon_{\theta}^t(x_t) + \sqrt{1-\alpha_t}\cdot s\nabla_{x_t} S(x_t)$, 
where $s$ is a hyperparameter controlling the amount of the guidance.
However, as also noted by \cite{AttendExcite}, vanilla post-hoc guidance may shift the domain of diffusion latents $x_{t-1}$ resulting in image quality degradations.
Indeed, according to the (deterministic) DDIM process (Eq. \ref{eq:ddim_deterministic_process}) after substituting $\epsilon_{\theta}^t(x_t)$ with $\hat{\epsilon}_{\theta}^t(x_t)$ we get
\begin{equation}
\begin{split}
\label{eq:shifting_gradient_term}
    x_{t-1} = 
    \sqrt{\alpha_{t-1}} \frac{x_t - \sqrt{1 - \alpha_t}\epsilon_{\theta}^t(x_t)}{\sqrt{\alpha_t}} + \\
    \sqrt{1 - \alpha_{t-1}}\epsilon_{\theta}^t(x_t) - \xi_t \nabla_{x_t} S(x_t), \\
    \xi_t = \sqrt{1-\alpha_t}\cdot s\left(\frac{\sqrt{1-\alpha_{t}}\sqrt{\alpha_{t-1}}}{\sqrt{\alpha_t}} - \sqrt{1-\alpha_{t-1}}\right),
\end{split}
\end{equation}
hence in Eq. \ref{eq:ddim_deterministic_process} we get the additional term $-\xi_t \nabla_{x_t} S(x_t)$ which may shift the original distribution of $x_{t-1}$.

To this end we introduce the \textit{Reweighting Attention Score Guidance (RASG)} strategy which benefits from the general DDIM backward process (Eq. \ref{eq:ddim_stochastic_process}) and introduces a gradient reweighting mechanism resulting in latent domain preservation.
Specifically, according to Claim \ref{prop:DDIM_sigma_proposition}, $x_{t-1}$ obtained either by Eq. \ref{eq:ddim_deterministic_process} or by Eq. 
\ref{eq:ddim_stochastic_process} will be in the required domain (see Fig. \ref{fig:RASG_domain_fig}). 
Hence if in Eq. \ref{eq:ddim_stochastic_process} we replace the stochastic component $\epsilon_t$ by the rescaled version of the gradient $\nabla_{x_t}S(x_t)$ (to make it closer to a sampling from $\mathcal{N}(\textbf{0},\textbf{1})$), we will keep $x_{t-1}$ in the required domain and at the same time will guide its sampling towards minimization of $S(x_t)$.
Rescaling of the gradient $\nabla_{x_t}S(x_t)$ is done by dividing it on its standard deviation (we do not change the mean to keep the direction of the $S(x_t)$ minimization, for more discussion see Appendix).
Thus, RASG sampling is done by the formula
\begin{equation}
\begin{aligned}
    x_{t-1} = \sqrt{\alpha_{t-1}} \frac{x_t - \sqrt{1 - \alpha_t}\epsilon_\theta(x_t)}{\sqrt{\alpha_t}} + \\
    \sqrt{1 - \alpha_{t-1} - \sigma_t ^ 2} \epsilon_\theta(x_t) +  \sigma_t \frac{\nabla_{x_t}S(x_t)}{\mbox{std}(\nabla_{x_t}S(x_t))}.
\end{aligned}
\end{equation}
Now let us define the function $S(x_t)$ (for more discussion on its choice see Appendix).
First we consider all cross-attention maps $A_{cross}$ with the output resolution of $\frac{H}{32}\times\frac{W}{32}$: $A_{cross}^1, \ldots, A_{cross}^m\in \mathbb{R}^{(H/32\cdot W/32)\times l}$, where $m$ is the number of such cross-attention layers, and $l$ is the number of token embeddings.
Then for each $k\in ind(\tau)\subset \{1,\ldots,l\}$ 
we average the attention maps and reshape to $\frac{H}{32}\times\frac{W}{32}$:
\begin{equation}
\begin{split}
    \overline{A}_{cross}^k(x_t) = \frac{1}{m}\sum_{i=1}^{m}A_{cross}^i[:,k] \in \mathbb{R}^{\frac{H}{32}\times \frac{W}{32}}.
\end{split}
\end{equation}
Using post-hoc guidance with $S(x_t)$ we aim to maximize the attention scores in the unknown region determined by the binary mask $M\in \{0,1\}^{H\times W}$, hence we take the average binary cross entropy between $\overline{A}^k(x_t)$ and $M$ ($M$ is downscaled with NN interpolation, $\sigma$ here is sigmoid):

\begin{equation}
\begin{aligned}
    S(x_t) = -\sum_{k\in ind(\tau)}\sum_{i=1}^{\frac{H}{32}\cdot \frac{W}{32}}[M_i\log \sigma(\overline{A}_{cross}^k(x_t)_i) + \\
    (1-M_i)\log (1- \sigma(\overline{A}_{cross}^k(x_t)_i))].
    \label{eq:defining_function_S}
\end{aligned}
\end{equation}

\subsection{Inpainting-Specialized Conditional Super-Resolution}
Here we discuss our method for high-resolution inpainting utilizing a pre-trained diffusion-based super-resolution model.
We leverage the fine-grained information from the known region to upscale the inpainted region (see Fig. \ref{figures:method_arch}.).
Recall that $I \in \mathbb{R}^{H \times W \times 3}$ is the original high-resolution image we want to inpaint, and $\mathcal{E}$ is the encoder of VQ-GAN \cite{VQGAN_paper}. 
We consider $X_0 = \mathcal{E}(I)$ and take a standard Gaussian noise $X_T\in  \mathbb{R}^{\frac{H}{4}\times \frac{W}{4}\times 4}$. 
Then we apply a backward diffusion process (Eq. \ref{eq:ddim_deterministic_process}) on $X_T$ by using the upscale-specialized SD model and conditioning it on the low resolution inpainted image $I^c_{low}$.
After each diffusion step we blend the estimated denoised latent $X_0^{pred} = (X_t - \sqrt{1-\alpha_t}\epsilon^t_{\theta}(X_t))/\sqrt{\alpha_t}$ with $X_0$ by using $M$:
\begin{equation}
    X_0^{pred} \leftarrow M\odot X_0^{pred} + (1-M)\odot X_0,
\end{equation}
and use the new $X_0^{pred}$ to determine the latent $X_{t-1}$ (by Eq. \ref{eq:ddim_deterministic_process}). After the last diffusion step $X_0^{pred}$ is decoded and blended (Poisson blending) with the original image $I$.

It's worth noting that our blending approach is inspired by seminal works \cite{DiffusionThermodynamics,avrahami2022blendeddiffusion} blending $X_t$ with the noisy latents of the forward diffusion.
In contrast, we blend high-frequencies from $X_0$ with the denoised prediction $X^{pred}_0$ allowing noise-free image details propagate from the known region to the missing one during all diffusion steps.

\section{Experiments}

\begin{table*}
    \centering
    \small
    \begin{tabular}{lcccc}
        \hline
        \textbf{Model Name} 
        & \textbf{\makecell{CLIP score}} $\uparrow$ 
        & \textbf{Accuracy} $\uparrow$ & \textbf{\makecell{Aesthetic score}} $\uparrow$ 
        & \textbf{\makecell{PickScore \\ \scriptsize (Ours vs \scriptsize Baselines)}} $\downarrow$ \\
        \hline
        GLIDE   \cite{GLIDE}                                        & 25.14 & 43.39 \% & 4.48 & 57.81 \% \\ 
        BLD  \cite{BlendedLatentDiffusion}                          & 24.23 & 49.12 \% & 4.81 & 55.35 \% \\
        SDXL Inpainting \cite{podell2023sdxl}                       & 24.79 & 53.94 \% & 4.69 & 58.69 \% \\
        Stable 2.0 Inpainting \cite{StableDiffusion}                    & 24.86 & 51.38 \% & 4.88 & 55.64 \% \\
        DreamShaper-ControlNet Inp. \cite{ControlNet}                     & 25.73 & 58.92 \% & 4.95 & 54.69 \% \\
        SmartBrush reprod. \cite{SmartBrush}                        & 25.79 & 66.36 \% & 4.85 & 54.23 \% \\
        DreamShaper Inpainting \cite{DreamShaper8Inp}                    & 25.62 & 59.02 \% & 4.96 & 51.98 \% \\
        \hline
        Ours                                       & \textbf{26.25} & \textbf{67.59} \% & \textbf{5.00} & \textbf{50.0} \% \\
        \hline
    \end{tabular}
    \caption{Quantitative comparison.}
    \label{tab:metrics}
\end{table*}

\begin{figure*}
    \centering
    \includegraphics[width=0.97\textwidth]{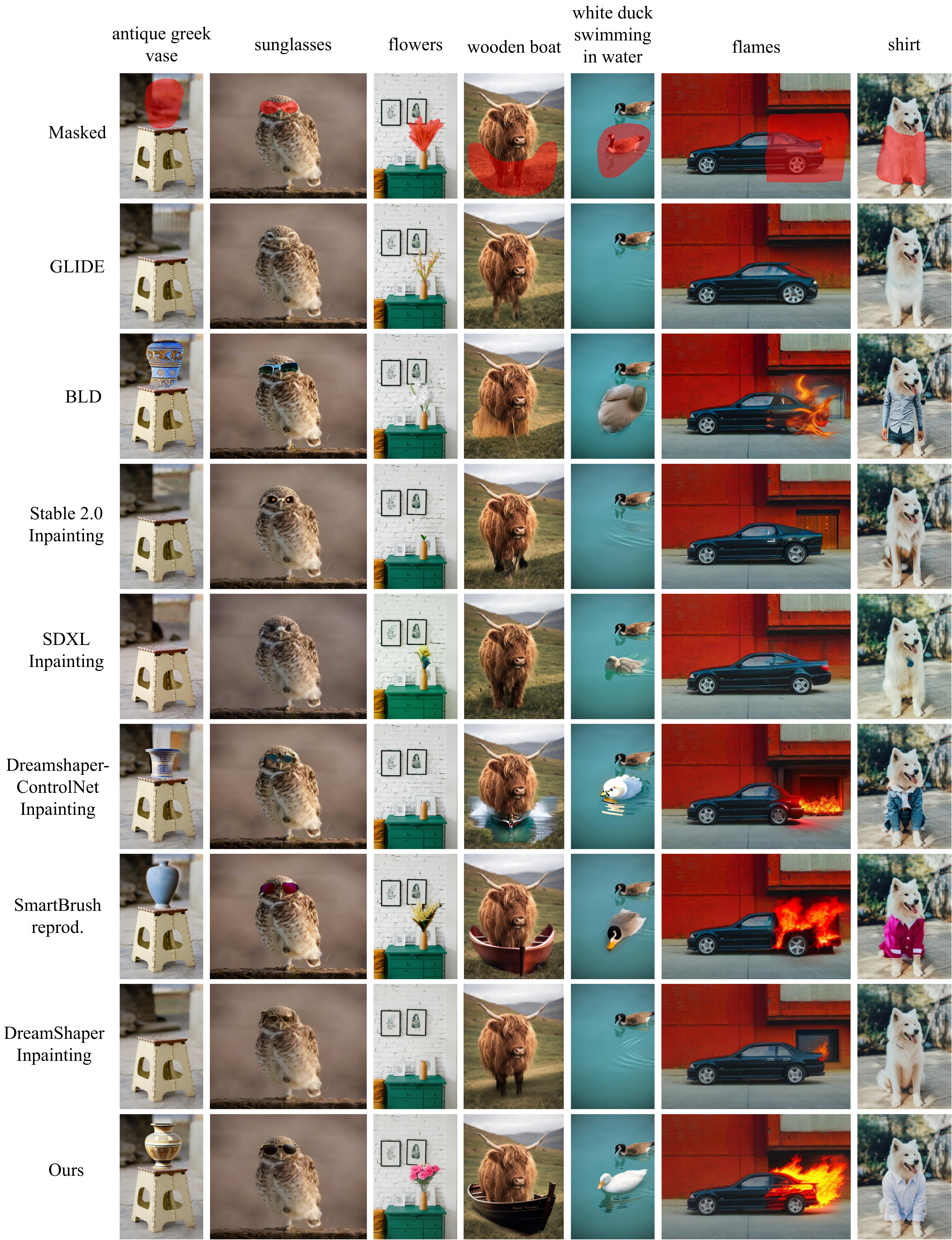}
    \caption{
    Comparison with state-of-the-art text-guided inpainting methods. Zoom in for details. For more comparison see Appendix.
    }
    \label{fig:qualitative_comparison}
\end{figure*}

\subsection{User Study}

\begin{figure*}
    \centering
    \includegraphics[width=0.8\linewidth]{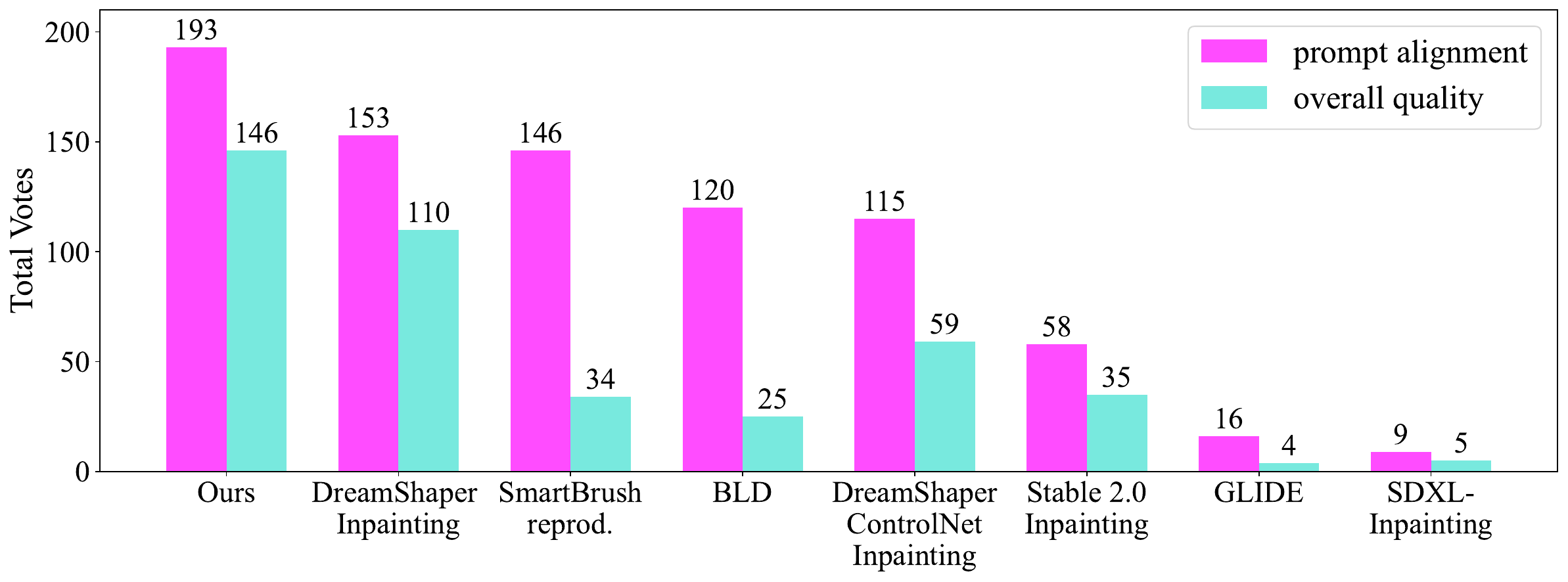}
    \caption{
    Total votes of each method based on our user study for \textit{prompt alignment} and \textit{overall quality}. Our method HD-Painter has a clear advantage over all competitors.
    }
    \label{fig:user_study}
\end{figure*}

We also performed a user study for a qualitative comparison with the competitor state-of-the art methods. The $12$ participants were shown $20$ \textit{(image, mask, prompt)}
triplets and the inpainting results of all methods in random order. For each sample image we asked to select the best results based on (\textit{i}) \textit{prompt alignment} and (\textit{ii}) \textit{overall quality},
allowing the choice of no methods when all methods were bad, 
or multiple methods when the quality was similar.
We calculate the total votes for all methods for each question. The results are presented in \cref{fig:user_study} demonstrating a clear advantage of our method in both aspects over all competitor methods.

\subsection{Implementation Details}

We use DreamShaper 8 \cite{DreamShaper8Inp} version of Stable Inpainting and Stable Super-Resolution 2.0 as image completion and inpainting-specialized super-resolution baselines respectively.
PAIntA is used to replace the self attention layers on the $H/32 \times W/32$ and $H/16 \times W/16$ resolutions for the first half of generation steps.
For RASG we select only cross-attention similarity matrices of the $H /32 \times W / 32$ resolution 
since utilizing higher resolutions did not offer significant improvements.

For hyperparameters $\{\sigma_t\}_{t=1}^T$ we chose 
$$
\sigma_t = \eta\sqrt{(1-\alpha_{t-1})/(1-\alpha_t)}\sqrt{1-\alpha_t/\alpha_{t-1}}, \; \eta = 0.1
$$

\subsection{Experimental Setup}

Here we compare with existing state-of-the-art methods such as GLIDE \cite{GLIDE}, Stable 2.0 Inpainting\footnote{\url{https://huggingface.co/stabilityai/stable-diffusion-2-inpainting}} \cite{StableDiffusion}, DreamShaper Inpainting \cite{DreamShaper8Inp}, Blended Latent Diffusion (BLD) \cite{BlendedLatentDiffusion}, ControlNet-Inpainting\footnote{\url{https://huggingface.co/lllyasviel/control_v11p_sd15_inpaint}} \cite{ControlNet} (with DreamShaper \footnote{\url{https://huggingface.co/Lykon/dreamshaper-8}} base), SDXL-Inpainting\footnote{\url{https://huggingface.co/spaces/diffusers/stable-diffusion-xl-inpainting}} \cite{podell2023sdxl} and SmartBrush \cite{SmartBrush}. As authors of the SmartBrush paper don't provide code and model, we reproduce it according to paper and refer to it as \textit{SmartBrush reprod.}. Specifically, we build SmartBrush reprod. based on DreamShaper text-to-image model in order to have fair comparison with our method, which uses DreamShaper Inpainting as a baseline.
We evaluate the methods on a random sample of 10000 (image, mask, prompt) triplets from the validation set of MSCOCO 2017 \cite{MSCOCO_data}, where the prompt is chosen as the label of the selected instance mask.
We noticed that when a precise mask of a recognizable shape is given to Stable Inpainting, it tends to ignore the prompt and inpaint based on the shape.
To prevent this, we use the convex hulls of the object segmentation masks and compute the metrics accordingly.

We evaluate the CLIP score on a cropped region of the image using the bounding box of the input mask. 
As CLIP score can still assign high scores to adversarial examples, we additionally compute the generation class accuracy.
So, we utilize a pre-trained instance detection model for MSCOCO: MMDetection \cite{chen2019mmdetection}.
We run it on the cropped area of the generated image, and, as there might be more than one objects included in the crop, we treat the example as positive if the prompt label is in the detected object list.

To measure the visual fidelity of the results we employ the LAION aesthetic score.\footnote{\url{https://github.com/christophschuhmann/improved-aesthetic-predictor}} The aesthetic score is computed by an MLP trained on 5000 image-rating pairs from the Simulacra Aesthetic Captions dataset \cite{SACDataset}, and can be used to assign a value from the $[0,10]$ range to images based on their aesthetic appeal.

Finally, we employ PickScore \cite{PickScore} as a combined metric of text-alignment and visual fidelity. 
Being trained on real user feedback PickScore is able to not only assess the prompt-faithfulness of inpainting methods but also the generation quality, while reflecting the complex requirements of users.
In our setting we apply PickScore between our vs other methods results and compute the percentage when it gives the advantage to our.

\subsection{Quantitative and Qualitative Analysis}

\Cref{tab:metrics} shows that our method outperforms the competitors in all metrics. It can be noticed that while SmartBrush trained over DreamShaper Inpainting improves the accuracy over the baseline, the CLIP score improvement is marginal and the overall quality is significantly dropped according to aesthetic score. On the other hand, our method significantly improves the prompt-alignment as measured by both CLIP score and accuracy while also maintaining the quality.

The examples in \cref{fig:qualitative_comparison} demonstrate qualitative comparison between our method and the other state-of-the-art approaches.
In many cases the baseline DreamShaper Inp. generates a background (\cref{fig:qualitative_comparison}, columns 1, 3, 5) or reconstructs the missing regions as continuation of the known region objects disregarding the prompt (\cref{fig:qualitative_comparison}, columns 4, 6, 7), while our method, thanks to the combination of PAIntA and RASG, successfully generates the target objects.
Notice that even though DreamShaper-ControlNet-Inpainting and SmartBrush reprod. may also generate the required object, the quality of the generation is poor compared to ours. 

Additionally, \cref{fig:teaser-image} demonstrates how effective our inpainting-specialized super-resolution is in seamlessly leveraging known region details for upscaling the generated region. 
We show more results in our Appendix, as well as comparison with vanilla Stable Super-Resolution approach \cite{StableDiffusion} used as an upscaling method after inpainting.

\subsection{Ablation Study}

In \cref{tab:ablation} we show that PAIntA and RASG separately on their own provide substantial improvements to the model quantitatively. 
We also provide more discussion on each of them in our supplementary material, including thorough analyses on their impact, demonstrated by visuals.

\begin{table}[h!]
        \centering
        \begin{adjustbox}{max width=\columnwidth}
        \begin{tabular}{lccc}
            \hline
            \textbf{Model Name} & \textbf{\makecell{CLIP \\ score}} $\uparrow$ & \textbf{Accuracy} $\uparrow$ & \textbf{\makecell{Aesthetic \\ score}} $\uparrow$ \\
            \hline    
            base (DreamShaper Inp.)             & 25.62 & 59.02 \% & 4.96 \\
            only RASG                            & 25.82 & 62.67 \% & 4.97\\
            only PAIntA                          & 26.06 & 64.17 \% & 4.99 \\
            \hline
            RASG \& PAIntA                       & \textbf{26.25} & \textbf{67.59} \% & \textbf{5.00} \\ 
            \hline
        \end{tabular} 
        \end{adjustbox}
        \caption{Ablation study for PAIntA and RASG.}
        \label{tab:ablation}
\end{table}

\vspace{-1mm}
\section{Conclusion}
\vspace{-1mm}
In this paper, we introduced a training-free approach to text-guided high-resollution image inpainting, addressing the prevalent challenges of prompt neglect: background and nearby object dominance. Our contributions, the Prompt-Aware Introverted Attention (PAIntA) layer and the Reweighting Attention Score Guidance (RASG) mechanism, effectively mitigate the mentioned issues leading our method to surpass the existing state-of-the-art approaches qualitatively and quantitatively. Additionally, our unique inpainting-specific super-resolution technique offers seamless completion in high-resolution images, distinguishing our method from existing solutions.

{
    \small
    \bibliographystyle{template/ieeenat_fullname}
    \bibliography{main}
}

\clearpage

\appendix
\renewcommand{\thesection}{Appendix \Alph{section}}

\section{Extended Qualitative Comparison}

\begin{figure*}
    \centering
    \includegraphics[width=0.95\textwidth]{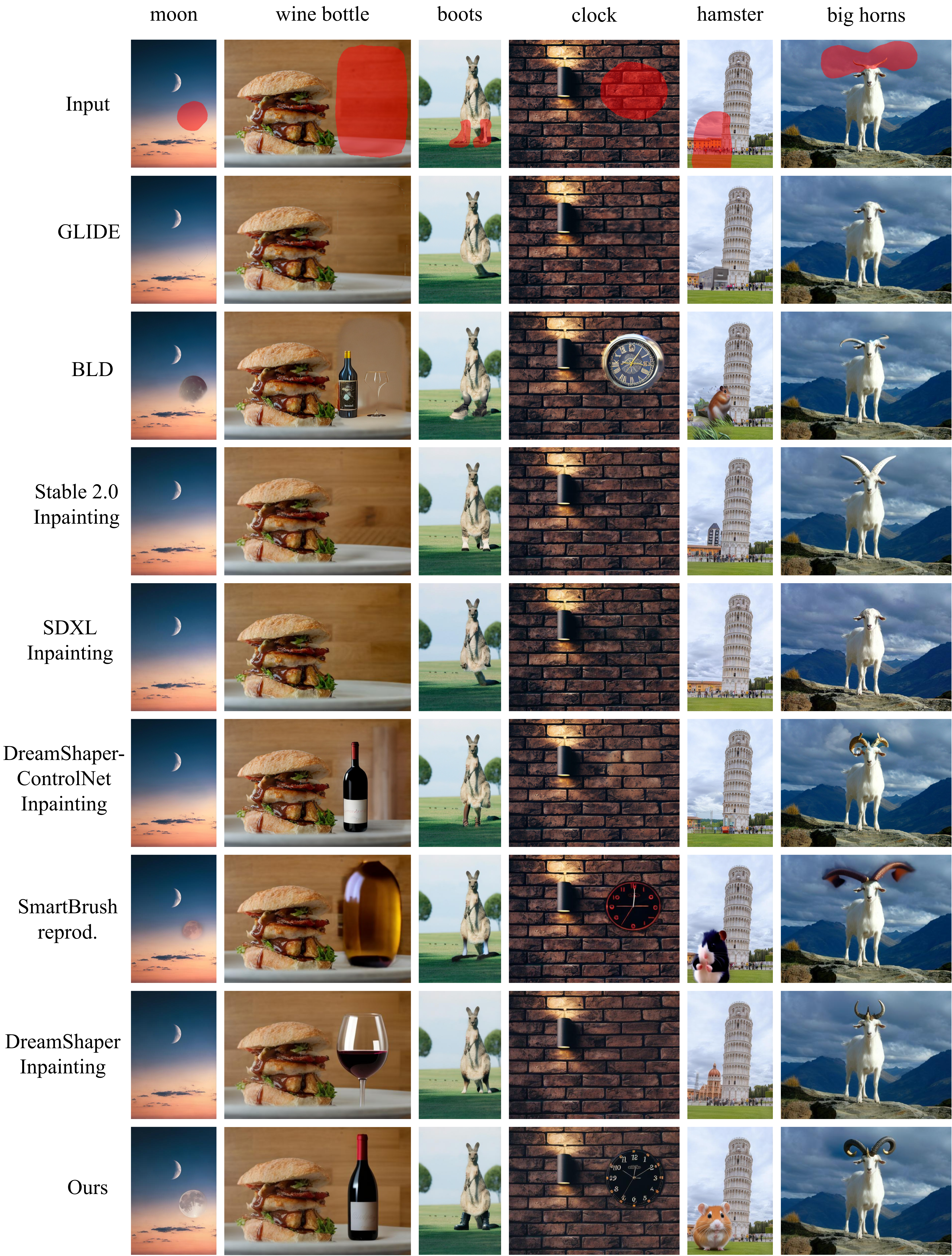}
    \caption{More qualitative comparison results. Zoom in to view high-resolution details.}
    \label{fig:appendix_qualitative_comparison}
\end{figure*}

\begin{figure*}
    \centering
    \includegraphics[width=\textwidth]{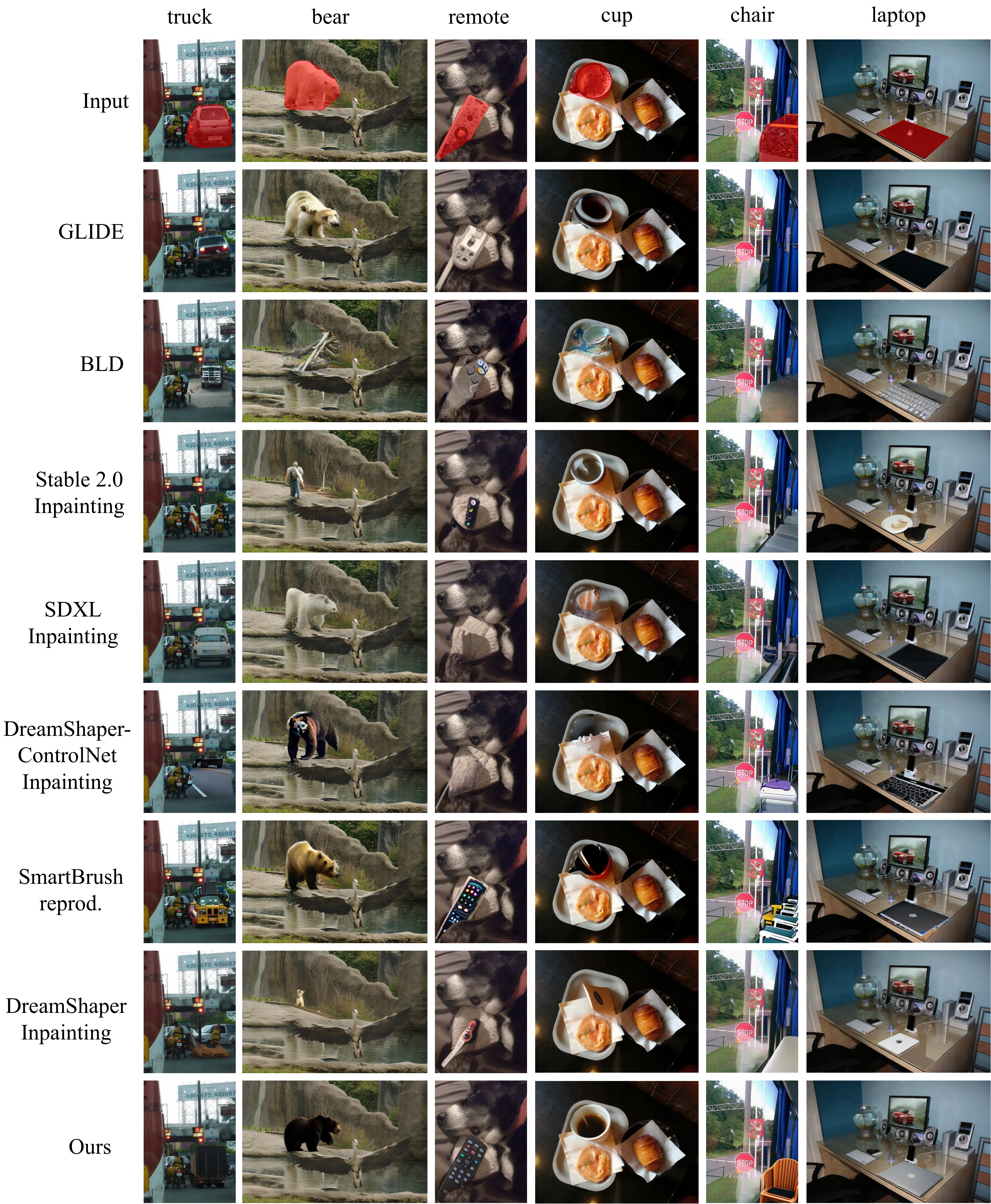}
    \caption{More qualitative comparison results on MSCOCO 2017.}
    \label{fig:appendix_coco_qualitative_comparison}
\end{figure*}

Here in Fig. \ref{fig:appendix_qualitative_comparison} we show more visual comparison with the other state-of-the-art methods. Fig. \ref{fig:appendix_coco_qualitative_comparison} includes more comparison on the validation set of MSCOCO 2017 \cite{MSCOCO_data}.
The results show the advantage of our method over the baselines.

\begin{figure*}
    \centering
    \includegraphics[width=\textwidth]{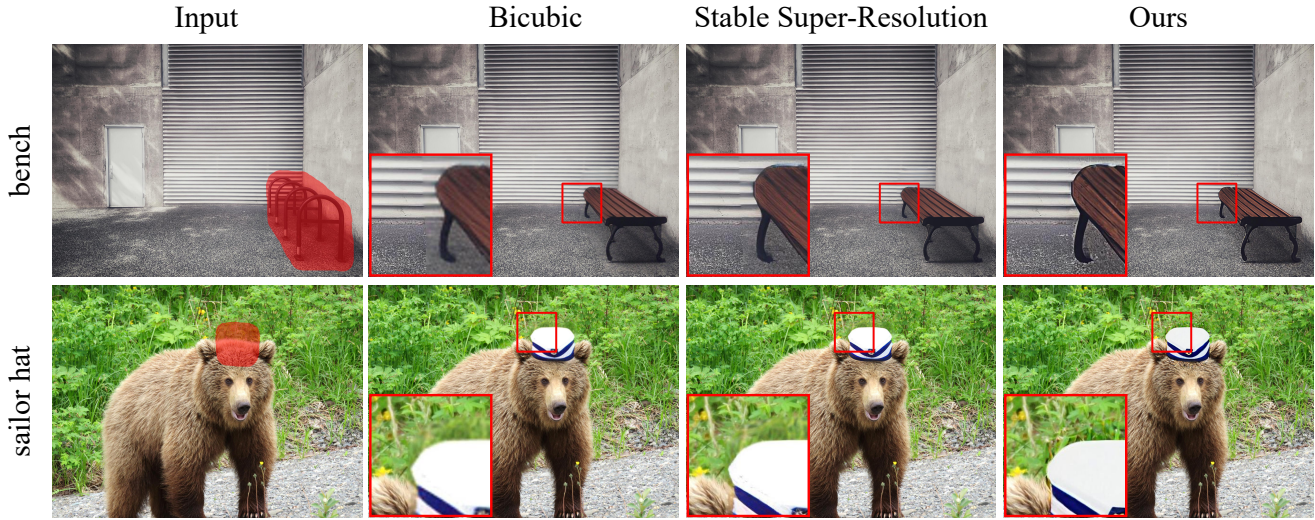}
    \captionof{figure}{Comparison of our inpainting-specialized super-resolution approach with vanilla upscaling methods for inpainting. Best viewed when zoomed in.}
    \label{fig:upscale-supp-bicubic_comparison}
    \vspace{-10pt}
\end{figure*}

In Fig. \ref{fig:upscale-supp-bicubic_comparison} we compare our inpainting-specialized super-resolution method with vanilla approaches of Bicubic or Stable Super-Resolution-based upscaling of the inpainting results followed by Poisson blending in the unknown region.
We can clearly see that our method, leveraging the known region fine-grained information, can seamlessly fill in with high quality.
In Figures \ref{fig:upscale-supp} and \ref{fig:upscale-supp2} we show more visual comparison between our method and the approach of Stable Super-Resolution.

\begin{figure*}[h!]
    \centering
    \includegraphics[width=0.95\textwidth]{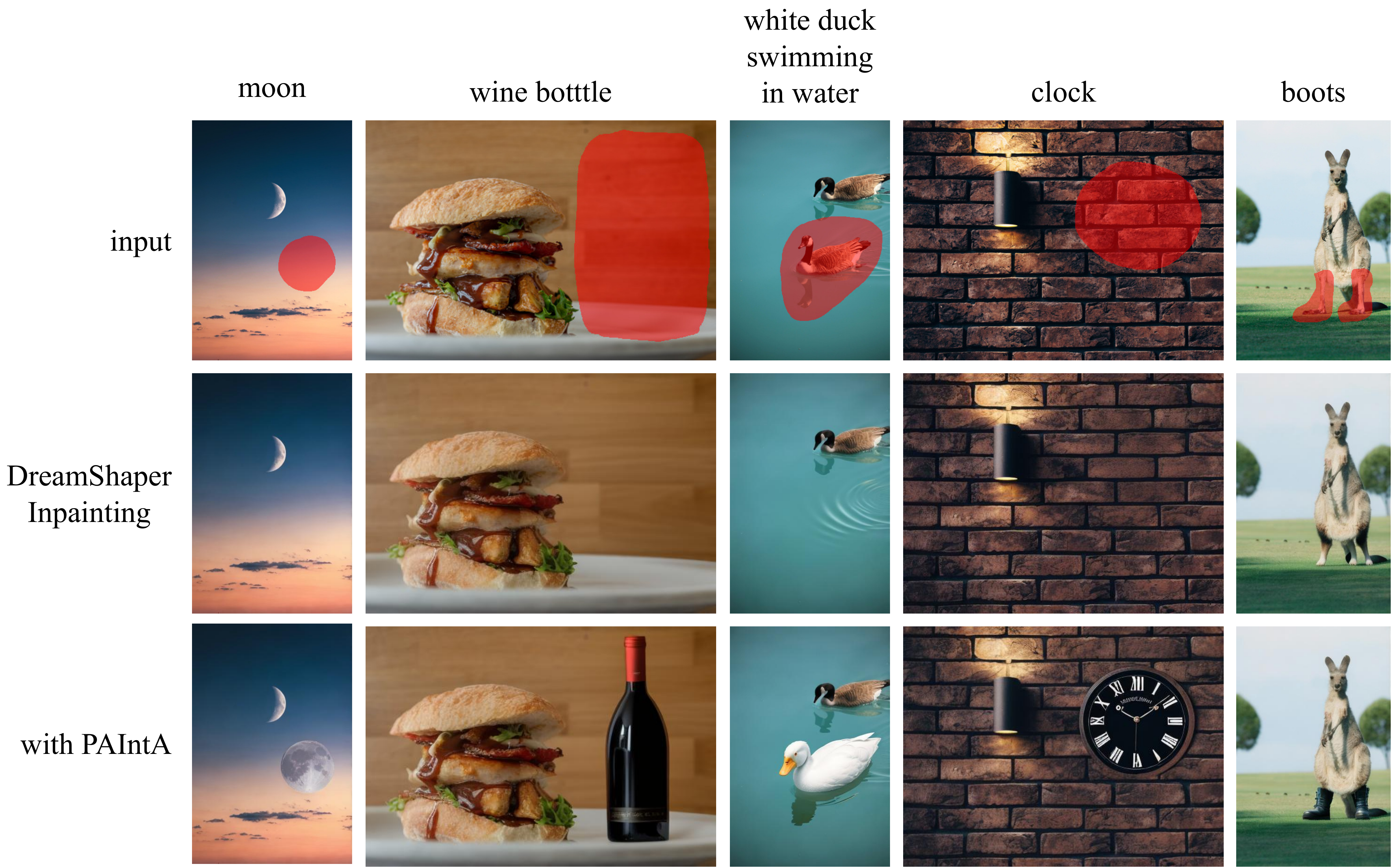}
    \caption{Visual ablation of PAIntA. Generated images use the same seed. In row 3 only PAIntA is used.} 
    \label{fig:appendix-painta-only}
\end{figure*}
\section{Discussion on PAIntA}

\begin{figure*}[h!]
    \centering
    \includegraphics[width=\textwidth]{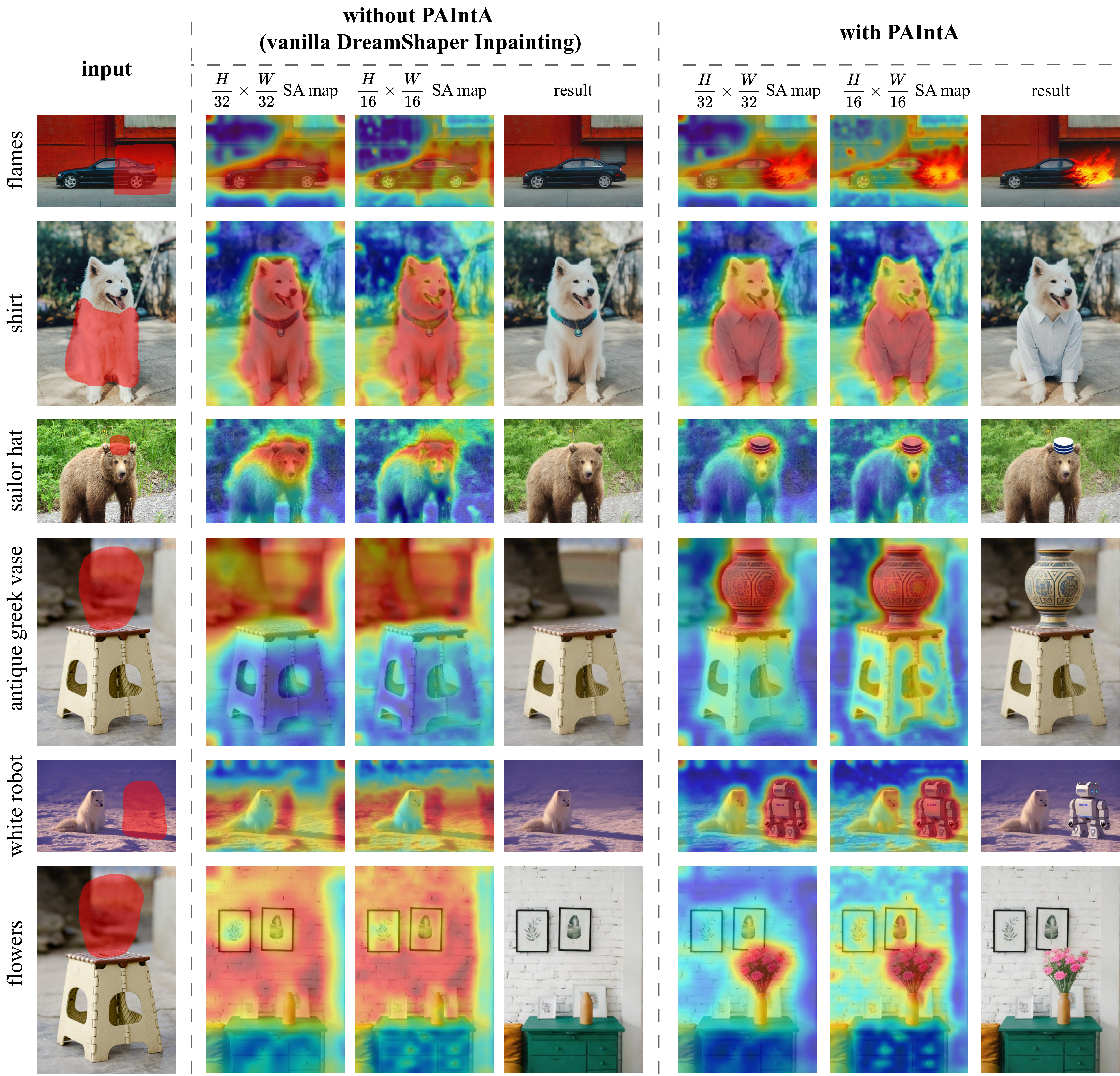}
    \caption{Comparison of self-attention similarity maps averaged across masked pixels for generations without/with PAIntA's scaling of the original self-attention scores. Images are generated from the same seed.} 
    \label{fig:appendix-self-attn-viz}
\end{figure*}

In this section we discuss the effectiveness of the proposed PAIntA module as a plug-in replacement for self-attention (SA) layers. To that end, first we visualize SA similarity maps averaged across masked locations from resolutions $H/16\times W/16$ and $H/32\times W/32$ where PAIntA is applied (see Fig. \ref{fig:appendix-self-attn-viz}). Then, we see that PAIntA successfully scales down the similarities of masked locations with prompt-unrelated locations from the known region, and, as a result, a prompt-specified object is generated inside the mask.

For a given resolution ($H/16\times W/16$ or $H/32\times W/32$), in order to visualize the average SA similarity map across masked pixels, first we resize the input mask to match the dimensions of the corresponding resolution (we use nearest interpolation in resize operation). Then, for each SA layer in the given resolution, we form a 2D similarity map by reshaping and averaging the similarity matrix rows corresponding to the masked region. Further, we average obtained 2D similarity maps across all SA layers (of the given resolution) and diffusion timesteps. 
More specifically, if $A^1_{self}, \ldots, A^L_{self} \in \mathbb{R}^{hw\times hw}$ ($h\times w$ is either $H/16\times W/16$ or $H/32\times W/32$) are the self-attention matrices of Stable Inpainting layers of the given resolution, and, respectively, are being updated by PAIntA to the matrices $\tilde{A^i}_{self}$ (see Eq. \ref{eq:self_attn_update}), 
then we consider the following similarity maps:
$$
    A = \frac{1}{|M|\cdot L}\;\sum_{i,M_i=1}\sum_{l=1}^L (A^l_{self})_i \in \mathbb{R}^{hw},
$$
$$
    \tilde{A} = \frac{1}{|M|\cdot L}\;\sum_{i,M_i=1}\sum_{l=1}^L (\tilde{A}^l_{self})_i \in \mathbb{R}^{hw},
$$
and reshape them to 2D matrices of size $h\times w$. 
So, $A_{ij}$ and $\tilde{A}_{ij}$ show the average amount in which masked pixels attend to to other locations in the cases of the vanilla self-attention and PAIntA respectively. 
Finally, in order to visualize the similarity maps, we use bicubic resize operation to match it with the image dimensions and plot the similarity heatmap using JET colormap from OpenCV \citep{itseez2015opencv}. 

Next, we compare the generation results and corresponding similarity maps obtained from above procedure when PAIntA's SA scaling is (the case of $\tilde{A}$) or is not (the case of $A$) used. Because PAIntA's scaling is only applied on $H/32\times W/32$ and $H/16\times W/16$ resolutions, we are interested in those similarity maps. Rows 1-3 in Fig. \ref{fig:appendix-self-attn-viz} demonstrate visualizations on \textit{nearby object dominance} issue (when known objects are continued to the inpainted region while ignoring the prompt) of the vanilla diffusion inpainting, while rows 4-6 demonstrate those of with \textit{background dominance} issue (when nothing is generated, just the background is coherently filled in).

For example, on row 1, Fig. \ref{fig:appendix-self-attn-viz} in case of \textit{DreamShaper Inpainting without PAIntA} generation, the average similarity of the masked region is dominated by the known regions of the car on both 16 and 32 resolutions. Whereas, as a result of PAIntA scaling application, the average similarity of the masked region with the car is effectively reduced, and the masked region is generated in accordance to the input prompt.

Row 4, Fig. \ref{fig:appendix-self-attn-viz} demonstrates an example where the result without PAIntA continues the background based on visual context instead of following the user prompt. In this case, visualization shows that usage of PAIntA successfully reduces the similarity of the masked region with the unrelated background. As a result, by reducing the similarity of masked region with the unrelated known regions PAIntA enables prompt-faithful generation.
You can find additional examples of PAIntA's effect on the final generation in \cref{fig:appendix-painta-only}.

\section{Discussion on RASG}

\begin{figure*}[h!]
    \centering
    \includegraphics[width=0.95\textwidth]{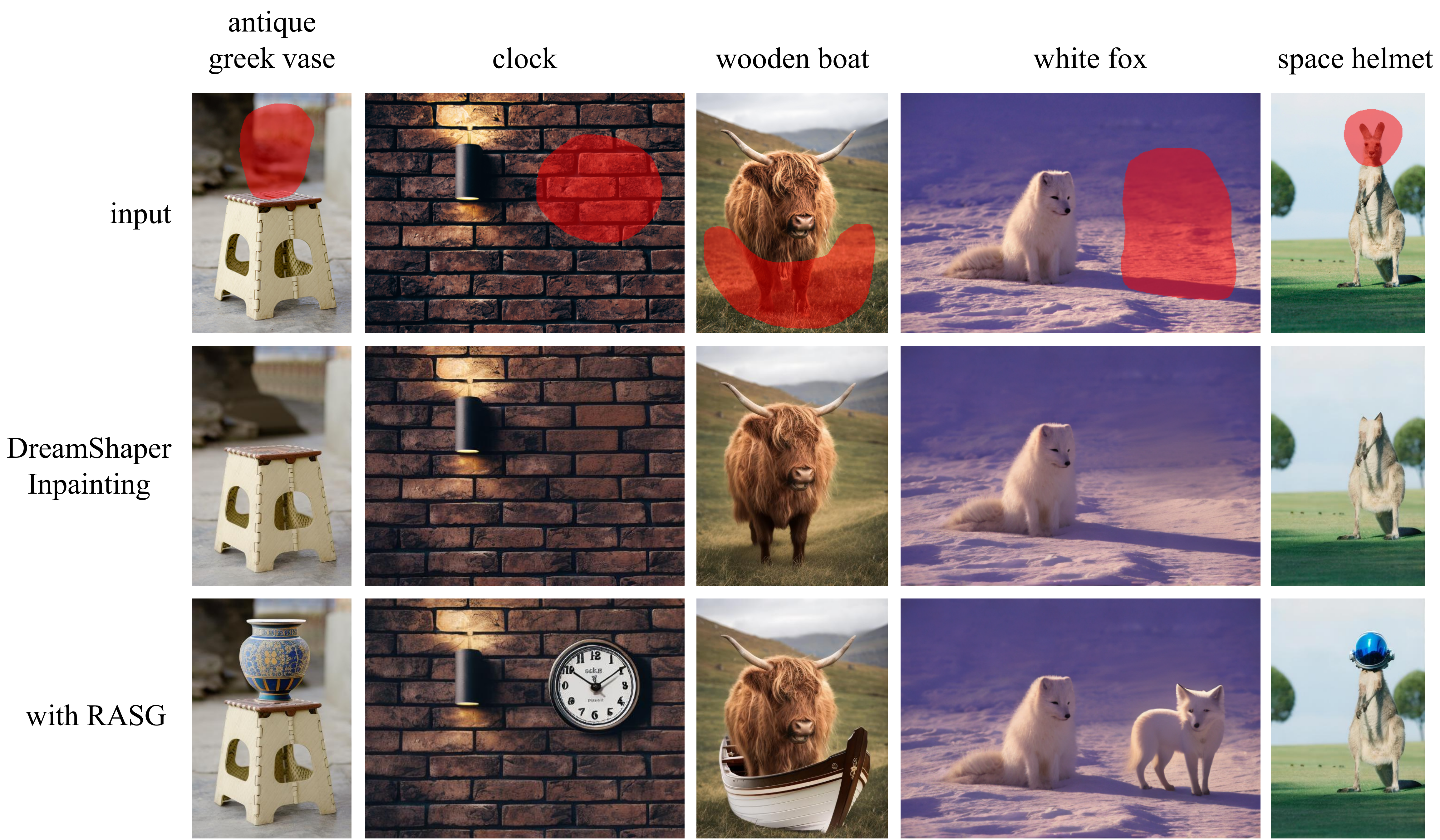}
    \caption{Visual ablation of RASG. Generated images use the same seed. In row 3 only RASG is used.} 
    \label{fig:appendix-rasg-only}
\end{figure*}

In this section we discuss the choice of RASG objective guidance function $S(x)$, then demonstrate the effect of RASG and motivate the part of gradient reweighting by its standard deviation.
Finally, we present additional examples of RASG's effect on the final generation in \cref{fig:appendix-rasg-only}.

\subsection{The Objective Function $S(x)$}
\begin{figure*}[h!]
    \centering
    \includegraphics[width=0.95\textwidth]{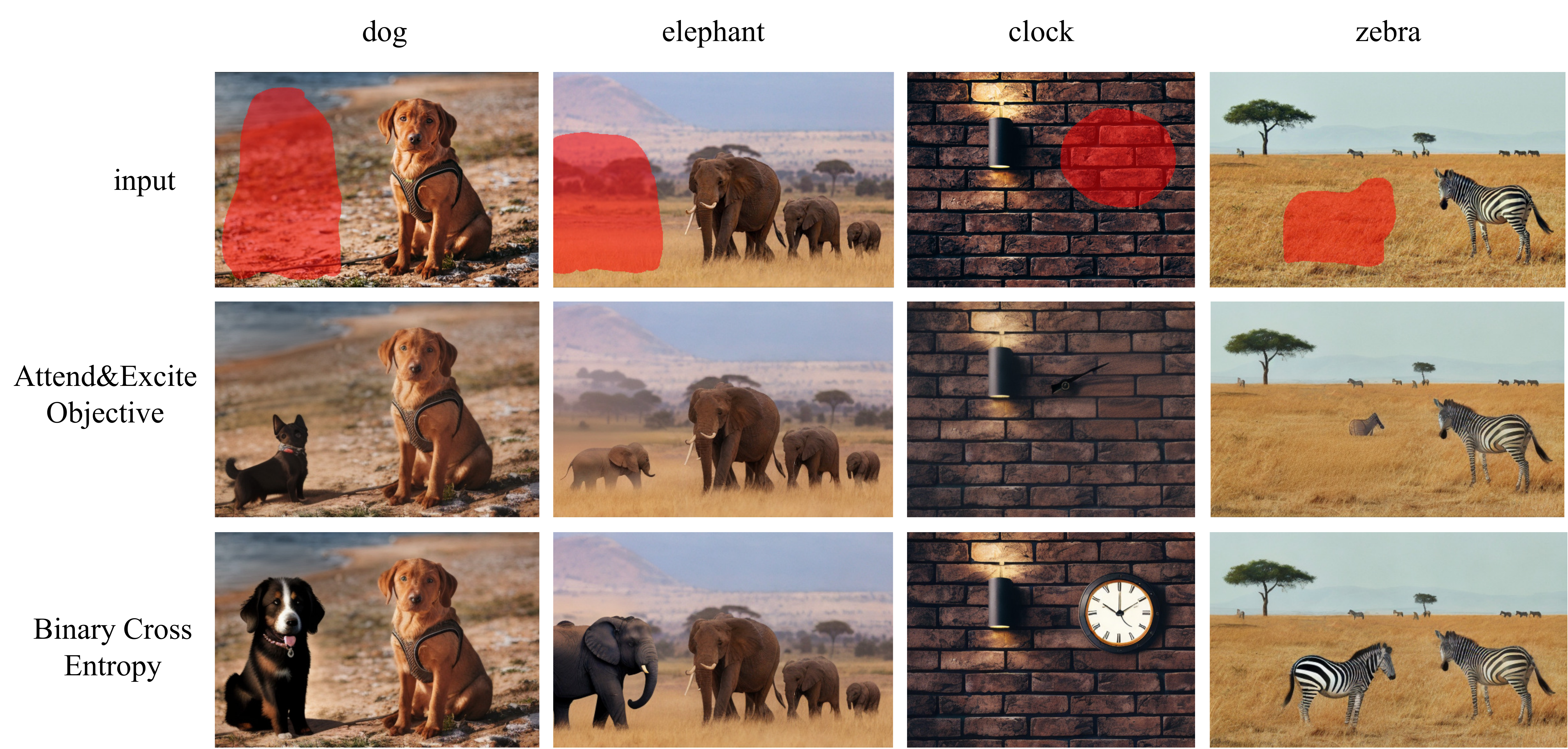}
    \caption{Comparison of the Binary Cross Entropy engery function to modifed version of Attend \& Excite. Images generated from the same seed.}
    \label{fig:sup:bce-ablation}
\end{figure*}

As we already mentioned in the main paper, Stable Inpainting may fail to generate certain objects in the prompt, completely neglecting them in the process. 
We categorized these cases into two types, namely background and nearby object dominance issues.
\cite{AttendExcite} also mentions these issues but for text-to-image generation task, and refers them as \textit{catastrophic neglect} problem.
To alleviate this problem \cite{AttendExcite} propose a mechanism called \textit{generative semantic nursing}, 
allowing the users to ``boost" certain tokens in the prompt, ensuring their generation.
In essence the mechanism is a post-hoc guidance with a chosen objective function maximizing the maximal cross-attention score of the image with the token which should be ``boosted".
This approach can be easily adapted to the inpainting task by just restricting the maximum to be taken in an unknown region so that the object is generated there, and averaging the objectives across all tokens, since we don't have specific tokens to ``boost", but rather care about all of them.
In other words, by our notations from the main paper, the following guidance objective funciton can be used:
\begin{equation}
    S(x_t) = -\frac{1}{|ind(\tau)|}\sum_{k\in ind(\tau)}\max_{i:\; M_i=1}\{\overline{A}^k(x_t)_i\}.
\end{equation}
However we noticed that with this approach the shapes/sizes of generated objects might not be sufficiently aligned with the shape/size of the input mask, 
which is often desirable for text-guided inpainting (see Fig. \ref{fig:sup:bce-ablation}).
Therefore, we utilize the segmentation property of cross-attention similarity maps, by so using \textit{Binary Cross Entropy} as the energy function for guidance (see Eq. \ref{eq:defining_function_S} in the main paper).
As can be noticed from Fig. \ref{fig:sup:bce-ablation} the results with the binary cross-entropy better fit the shape of the inpaining mask.

\subsection{Effect of RASG Strategy}

\begin{figure*}
    \centering
    \includegraphics[width=0.95\textwidth]{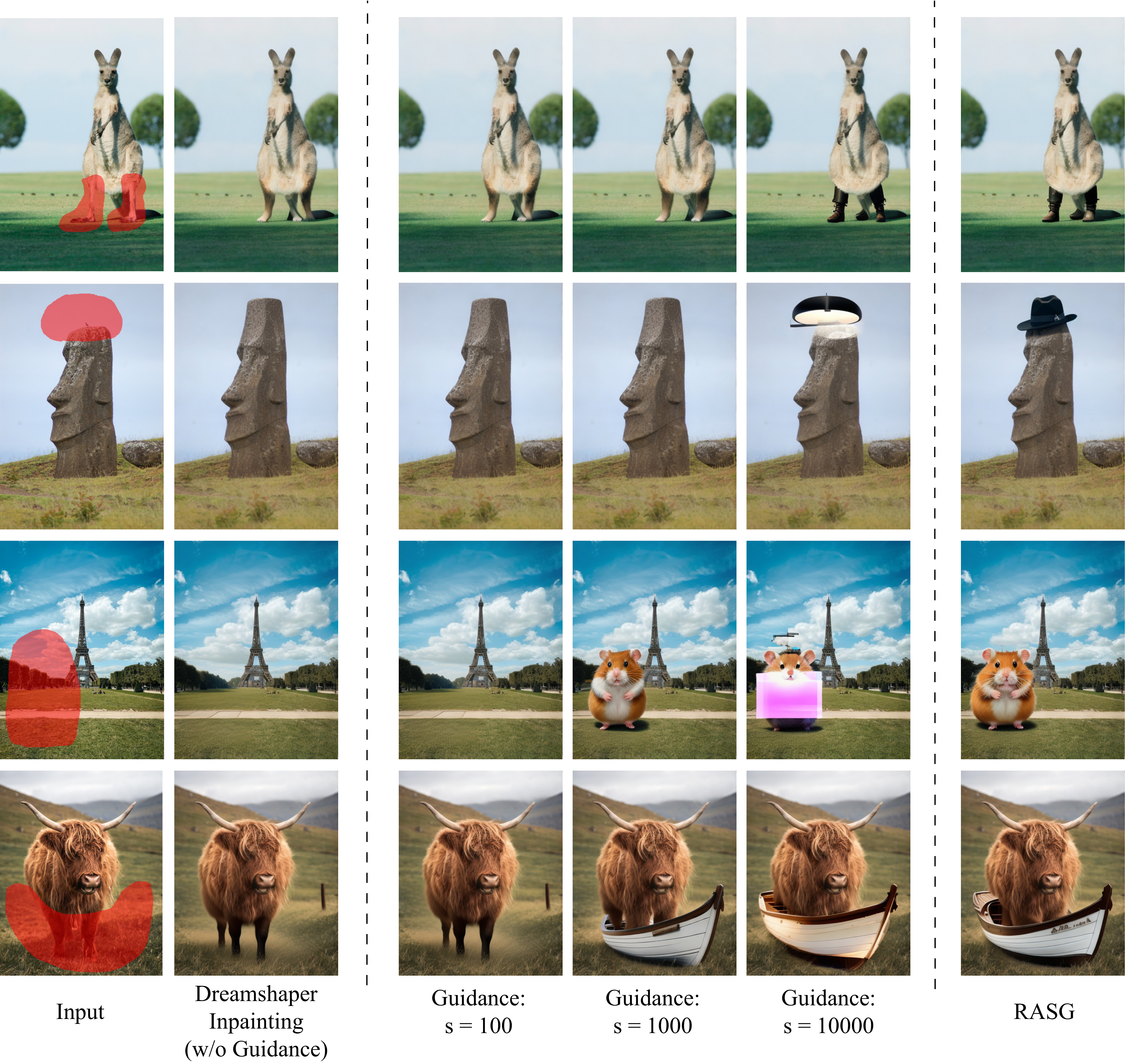}
    \caption{Comparison of RASG strategy with default Stable Inpainting and vanilla guidance mechanism with different guidance scales. In contrast to vanilla guidance, where the generation highly depends on the guidance scale, RASG consistently produces naturally looking and prompt-aligned results.}
    \label{fig:sup:rasg-examples}
\end{figure*}

Although the objective function $S(x)$ defined by Eq. \ref{eq:defining_function_S} (main paper) results in better mask shape/size aligned inpainting, the vanilla post-hoc guidance may lead the latents to become out of their trained domain as also noted by \cite{AttendExcite}: \textit{``many updates of} $x_t$ \textit{may lead to the latent becoming out-of-distribution, resulting in incoherent images"}.
Due to this the post-hoc guidance mechanism (semantic nursing) by \cite{AttendExcite} is done using multiple iterations of very small, iterative perturbations of $x_t$,
which makes the process considerably slow.
In addition, the generation can still fail if the iterative process exceeds the maximum iteration limit without reaching the necessary thresholds.

Thanks to RASG's seamless integration of the $\nabla_{x_t}S(x_t)$ gradient component into the general form of DDIM diffusion sampling, our RASG mechanism keeps the modified latents $x_t$ within the expected distribution, while introducing large enough perturbations to $x_t$ with only one iteration of guidance per time-step.
This allows to generate the objects described in the prompts coherently with the known region without extra-cost of time.

Fig. \ref{fig:sup:rasg-examples} demonstrates the advantage of RASG's strategy over the vanilla guidance mechanism.
Indeed, in the vanilla post-hoc guidance there is a hyperparameter $s$ controlling the amount of guidance. 
When $s$ is too small (e.g. close to $0$ or for some cases $s=100$) the vanilla guidance mechanism does not show much effect due to too small guidance from $s\nabla_{x_t}S(x_t)$.
Then with increasing the hyperparameter ($s=1000, 10000$) one can notice more and more text/shape alignment with prompt/inpainting mask, however the generated results are unnatural and incoherent with the known region.
This is made particularly challenging by the fact, that different images, or even different starting seeds with the same input image
might require different values of the perturbation strength to achieve the best result.
In contrast, RASG approach is \textit{hyperparameter-free} allowing both: prompt/mask-aligned and naturally looking results.

\subsection{Rescaling with Standard Deviation}

\begin{figure*}[h!]
    \centering
    \includegraphics[width=0.95\textwidth]{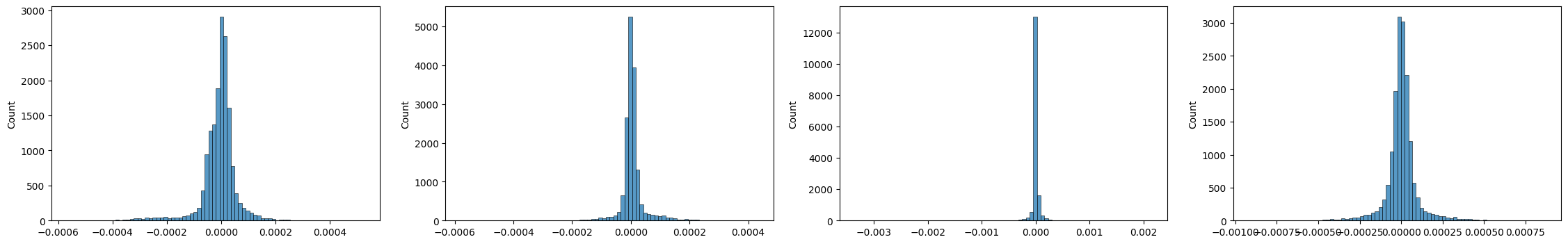}
    \caption{Histogram of $\nabla_{x_t} S(x_t)$  values (i.e. before gradient standardization)}
    \label{fig:sup:rasg-hist-before}
\end{figure*}

\begin{figure*}[h!]
    \centering
    \includegraphics[width=0.95\textwidth]{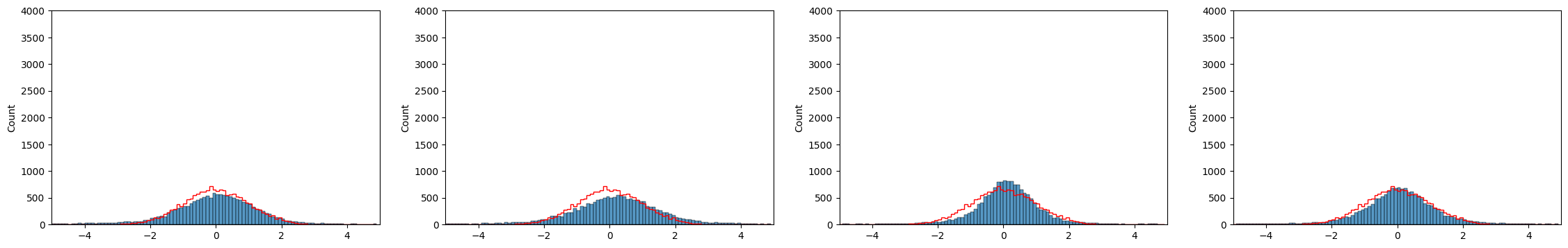}
    \caption{Histogram of $\frac{\nabla_{x_t} S(x_t)}{std(\nabla_{x_t} S(x_t))}$ values (i.e. after gradient standardization)}
    \label{fig:sup:rasg-hist-after}
\end{figure*}

The core idea of RASG is to automatically scale perturbation using certain heuristics,
such that the guidance process has a consistent effect on the output, without harming the quality of the image.
Our main heuristic relies on the fact that  \cite{DDIM} have defined a parametric family of
stochastic denoising processes, which can all be trained using the same training objective as DDPM \cite{DDPM}.
Recall the general form of parametric family of DDIM sampling processes:
\begin{equation}
\begin{aligned}
\label{eq:sup:ddim_stochastic_process}
    x_{t-1} = \sqrt{\alpha_{t-1}} \frac{x_t - \sqrt{1 - \alpha_t}\epsilon^t_\theta(x_t)}{\sqrt{\alpha_t}} + \\
    \sqrt{1 - \alpha_{t-1} - \sigma_t ^ 2} \epsilon^t_\theta(x_t) +  \sigma_t \epsilon_t,
\end{aligned}
\end{equation}
where $\epsilon_t\sim\mathcal{N}(\textbf{0},\textbf{1})$.
Particularly $\epsilon_t$ can be taken to be collinear with the gradient $\nabla_{x_t}S(x_t)$ which will result in $x_{t-1}$ distribution preservation by at the same time guiding the generation process towards minimization of $S(x_t)$.

Therefore we propose to scale the gradient $\nabla_{x_t}S(x_t)$ with a value $\lambda$ and use instead of $\epsilon_t$ in the general form of DDIM.
To determine $\lambda$ we analyse the distribution of $\nabla_{x_t}S(x_t)$ and found out that the values of the gradients have a distribution very close 
to a gaussian distribution, with $0$ mean and some arbitary $\sigma$, which changes over time-step/image (\cref{fig:sup:rasg-hist-before}).
Therefore, computing the standard deviation of the values of $\nabla_{x_t} S(x_t)$, and normalizing it by $\lambda = \frac{1}{std(\nabla_{x_t}S(x_t))}$
results in the standard normal distribution (see Fig. \ref{fig:sup:rasg-hist-after}).
So the final form of RASG guidance strategy is
\begin{equation}
\begin{aligned}
    x_{t-1} = \sqrt{\alpha_{t-1}} \frac{x_t - \sqrt{1 - \alpha_t}\epsilon^t_\theta(x_t)}{\sqrt{\alpha_t}} + \\ 
    \sqrt{1 - \alpha_{t-1} - \sigma_t ^ 2} \epsilon^t_\theta(x_t) +  \sigma_t \frac{\nabla_{x_t}S(x_t)}{std(\nabla_{x_t}S(x_t))}.
\end{aligned}
\end{equation}

\section{Limitations}

\begin{figure}[h!]
    \centering
    \includegraphics[width=\linewidth]{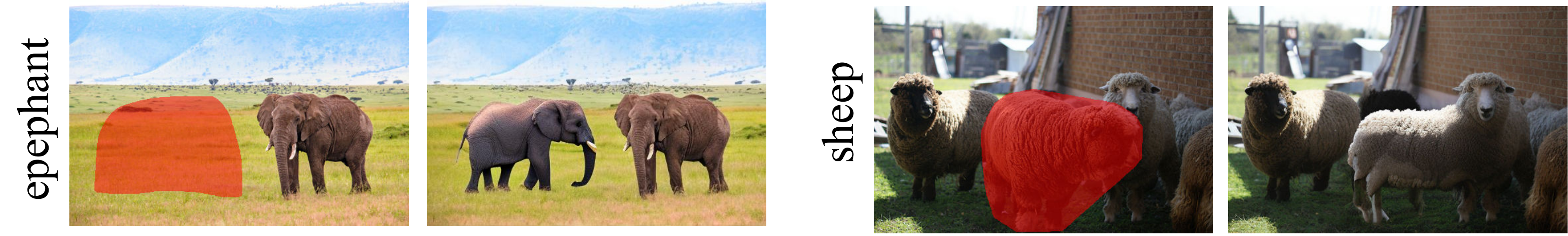}
    \caption{Failure examples produced by our approach.} 
    \label{fig:failure-cases}
\end{figure}

Although our method improves the prompt-alignment of existing text-guided inpainting approaches, it still has a dependency on the backbone model, hence inherits some quality limitations. 
Particularly it may generate extra limbs (the elephant in Fig. \ref{fig:failure-cases} has $5$ legs) or illogical appearances (the sheep appears to have two bodies in Fig. \ref{fig:failure-cases} after the inpainting).

\section{Potential negative impacts}

Our research strives to enhance the accuracy of object generation within the scope of text-guided image inpainting. However, it is crucial to acknowledge the potential negative impacts. The technology could be exploited to create deceptive imagery or disseminate misinformation, raising ethical concerns. While our method is training-free and does not introduce new biases, it is imperative to consider the potential propagation of biases from the base models we build upon. These biases could lead to the generation of content that inadvertently reflects societal or historical prejudices.

To counter these issues, it is essential for the broader research community to establish ethical standards and develop robust methods to detect AI-generated content. Furthermore, efforts should be made to diversify training datasets to reduce inherent biases. While these challenges are significant, the positive implications of our work in areas such as creative arts, design and content creation, when used responsibly, have the potential to surpass the negative repercussions.
\section{More Examples of Our Method}

\begin{figure*}
    \centering
    \includegraphics[width=\textwidth]{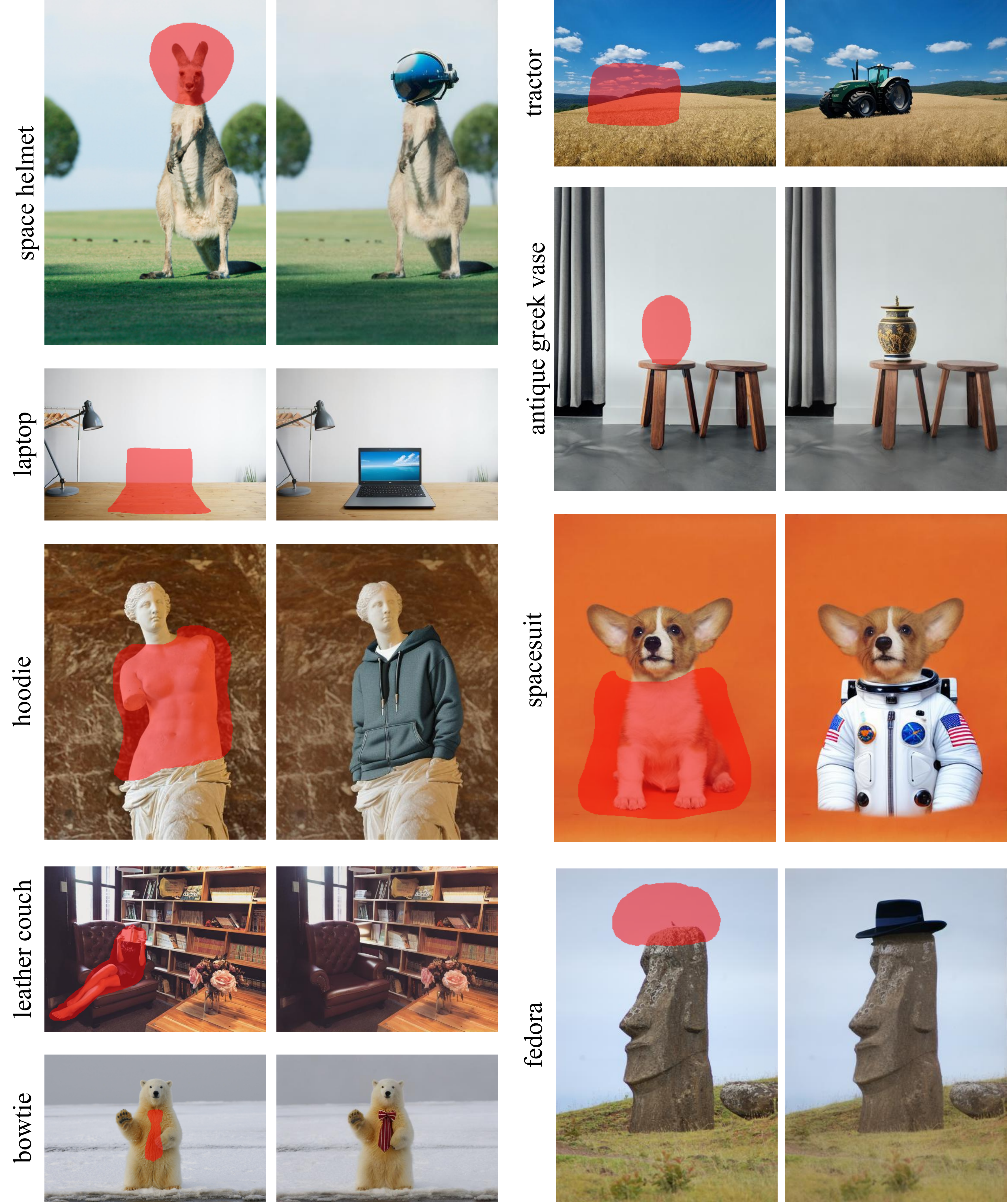}
    \caption{More results of our method.}
    \label{fig:appendix-gen-examples-1}
\end{figure*}

\begin{figure*}
    \centering
    \includegraphics[width=0.9\textwidth]{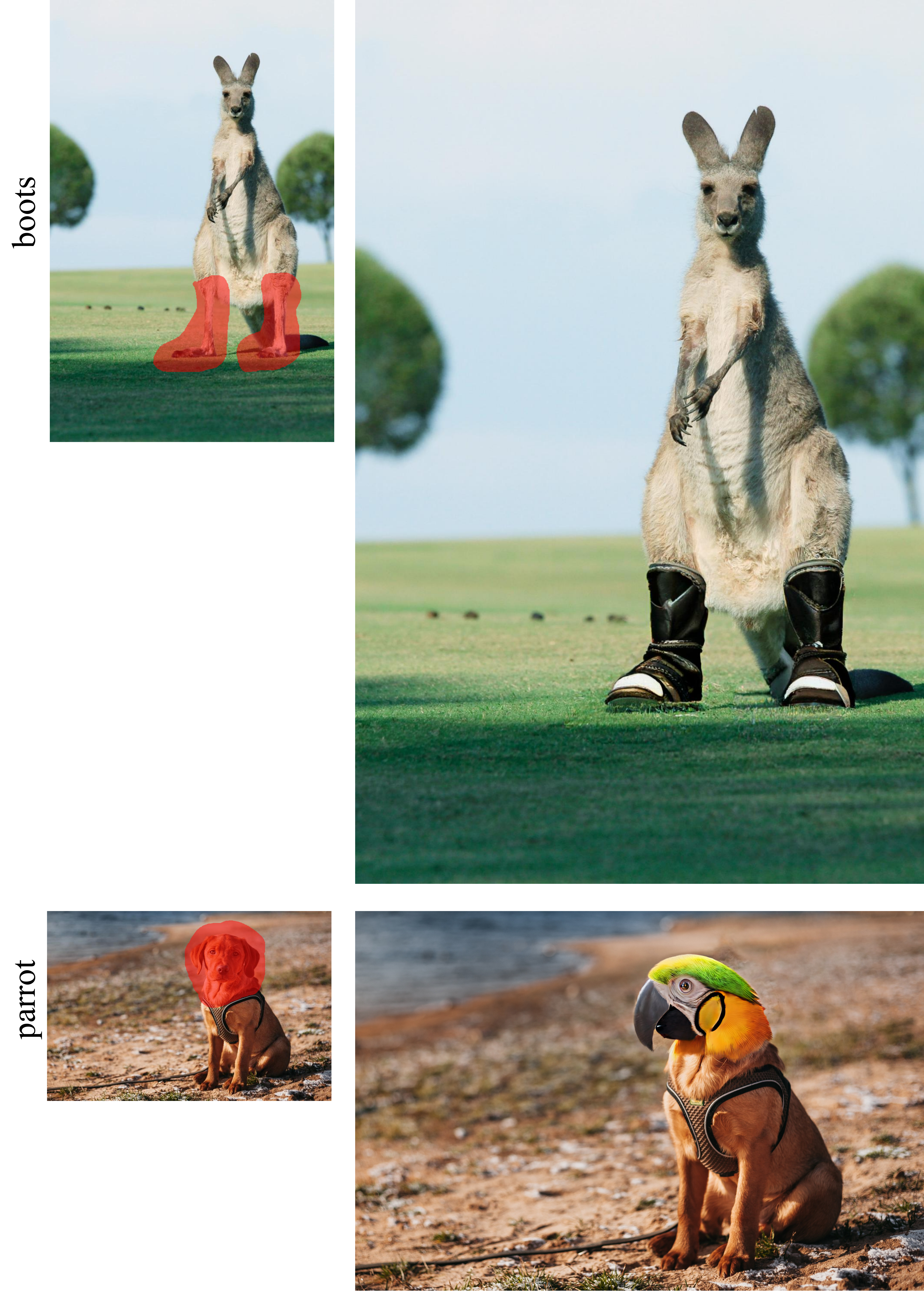}
    \caption{More high-resolution results of our method. Zoom in to view high-resolution details.}
    \label{fig:appendix-gen-examples-2}
\end{figure*}

\begin{figure*}
    \centering
    \includegraphics[width=0.9\textwidth]{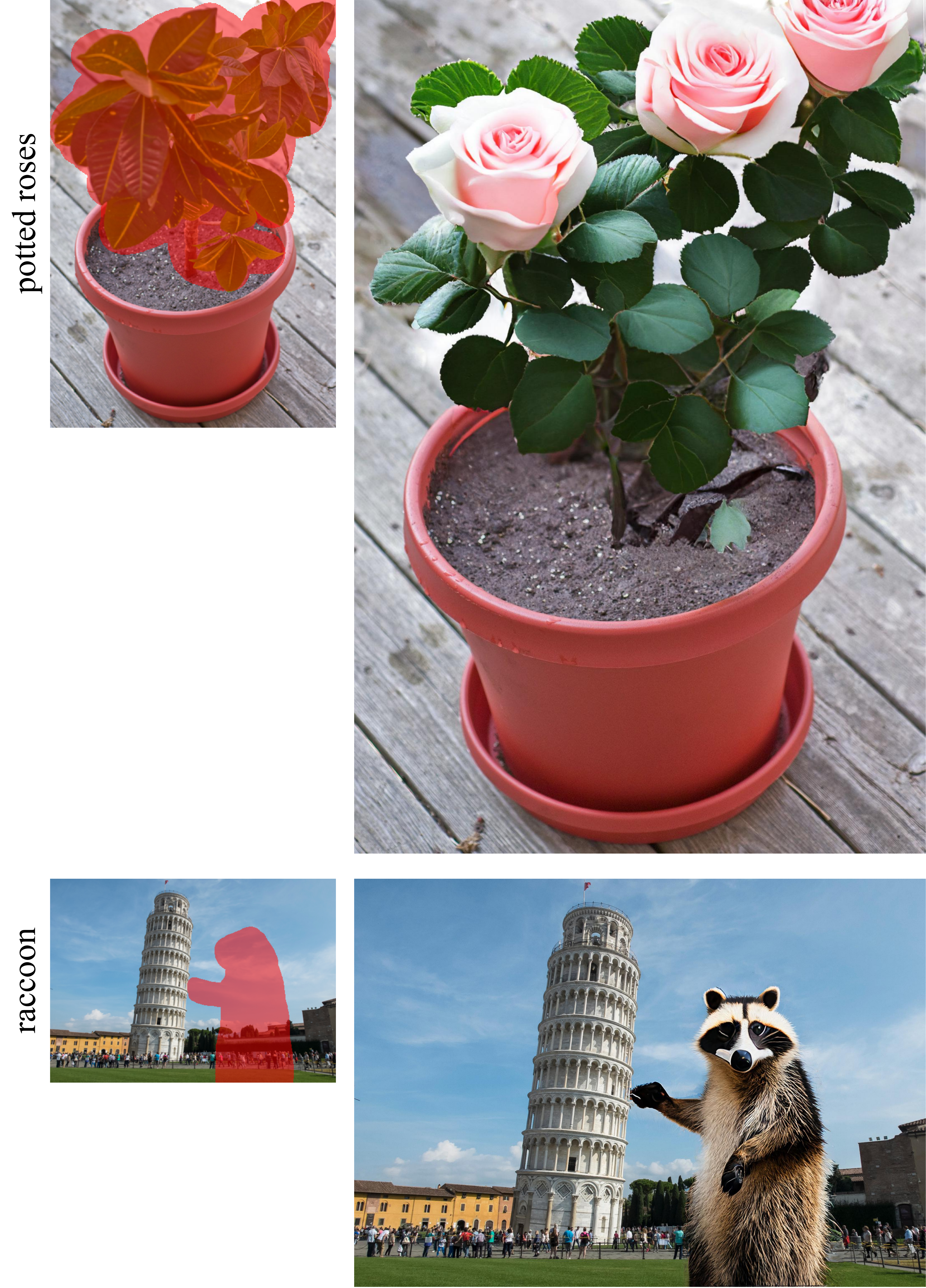}
    \caption{More high-resolution results of our method. Zoom in to view high-resolution details.}
    \label{fig:appendix-gen-examples-3}
\end{figure*}

\begin{figure*}
    \centering
    \includegraphics[width=0.9\textwidth]{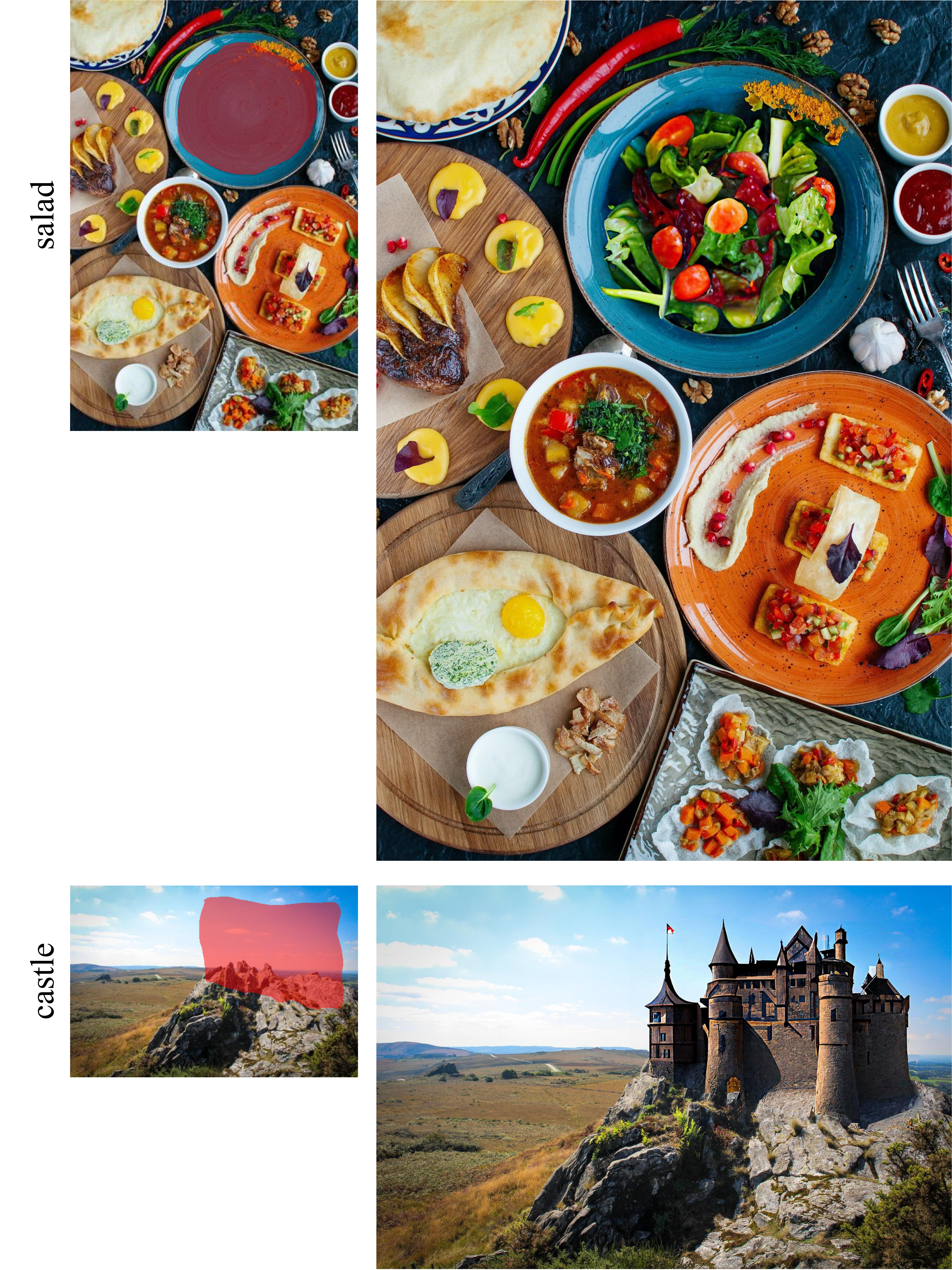}
    \caption{More high-resolution results of our method. Zoom in to view high-resolution details.}
    \label{fig:appendix-gen-examples-4}
\end{figure*}

We present more results of our method both for low-resolution (512 for the long side) images (Fig. \ref{fig:appendix-gen-examples-1}),
as well as high-resoltuion (2048 for the long side) (Figures \ref{fig:appendix-gen-examples-2}, \ref{fig:appendix-gen-examples-3}, \ref{fig:appendix-gen-examples-4}). 

\clearpage


\begin{figure*}
    \centering
    \includegraphics[width=0.8\linewidth]{appendix/figures/superres_supp.pdf}
    \captionof{figure}{Comparison between vanilla SD 2.0 upscale and our approach. In all examples the large side is 2048px. The cropped region is 256x256px. Best viewed when zoomed in.}
    \label{fig:upscale-supp}
\end{figure*}

\begin{figure*}
    \centering
    \includegraphics[width=0.85\linewidth]{appendix/figures/superres_supp2.pdf}
    \captionof{figure}{Comparison between vanilla SD 2.0 upscale and our approach. In all examples the large side is 2048px. The cropped region is 256x256px. Best viewed when zoomed in.}
    \label{fig:upscale-supp2}
\end{figure*}

\newcommand{\tabsizeEO}{0.1666\textwidth}

\end{document}